\documentclass[acmsmall,screen,nonacm]{acmart}

\usepackage{amsthm}
\usepackage{amsmath}
\usepackage{caption}
\usepackage{bussproofs}
\usepackage[linesnumbered,vlined, commentsnumbered, figure, noend]{algorithm2e}
\usepackage{multirow}
\usepackage[framemethod=TikZ]{mdframed}
\usepackage{xcolor}
\usepackage{multicol}

\definecolor{lightgray}{rgb}{0.95,0.95,0.95}
\makeatletter
\newtheorem*{rep@theorem}{\rep@title}
\newcommand{\newreptheorem}[2]{%
\newenvironment{rep#1}[1]{%
 \def\rep@title{#2~\ref{##1}}%
 \begin{rep@theorem}}%
 {\end{rep@theorem}}}
\makeatother

\makeatletter
\newtheorem*{rep@lemma}{\rep@title}
\newcommand{\newreplemma}[2]{%
\newenvironment{rep#1}[1]{%
 \def\rep@title{#2~\ref{##1}}%
 \begin{rep@lemma}}%
 {\end{rep@lemma}}}
\makeatother

\newtheorem{definition}{Definition}
\newtheorem{theorem}{Theorem}
\newtheorem{lemma}{Lemma}

\newreptheorem{theorem}{Theorem}
\newreplemma{lemma}{Lemma}

\setcopyright{none}
\acmYear{2023}
\usepackage{booktabs}   %% For formal tables:
\usepackage{subcaption} %% For complex figures with subfigures/subcaptions
\usepackage{graphicx}
\newcommand{\tool}{\textsf{GNN-Infer}}

\newcounter{todocounter}

\begin{document}

%% Title information
\title{Inferring Properties of Graph Neural Network}

%% Author information
%% Contents and number of authors suppressed with 'anonymous'.
%% Each author should be introduced by \author, followed by
%% \authornote (optional), \orcid (optional), \affiliation, and
%% \email.
%% An author may have multiple affiliations and/or emails; repeat the
%% appropriate command.
%% Many elements are not rendered, but should be provided for metadata
%% extraction tools.

%% Author with single affiliation.

\author{Thanh-Dat Nguyen}
\email{thanhdatn@student.unimelb.edu.au}
\affiliation{%
  \institution{University of Melbourne}
%  \city{Melbourne}
%   \state{Ohio}
  \country{Australia}
}

\author{Minh-Hieu Vu}
\affiliation{%
  \institution{Independent Researcher}
%  \city{Hanoi}
%   \state{Ohio}
  \country{Vietnam}
}

\author{Thanh Le-Cong}
\affiliation{%
  \institution{University of Melbourne}
%  \city{Hanoi}
%   \state{Ohio}
  \country{Australia}
}

\author{Xuan Bach D. Le}
\email{bach.le@unimelb.edu.au}
\affiliation{%
  \institution{University of Melbourne}
%  \city{Melbourne}
%   \state{Ohio}
  \country{Australia}
}

\author{David Lo}
\email{davidlo@smu.edu.sg}
\affiliation{%
  \institution{Singapore Management University}
%  \city{Melbourne}
%   \state{Ohio}
  \country{Singapore}
}

\author{ThanhVu Nguyen}
\email{tvn@gmu.edu}
\affiliation{%
  \institution{George Mason University}
%  \city{Fairfax}
%   \state{Ohio}
  \country{USA}
}

\author{Corina Pasareanu}
\email{pcorina@andrew.cmu.edu}
\affiliation{%
  \institution{Carnegie Mellon University}
%  \city{Fairfax}
%   \state{Ohio}
  \country{USA}
}

% \author{David Lo}
% \email{davidlo@smu.edu.sg}
% \affiliation{%
%   \institution{Singapore Management University}
% %  \city{Melbourne}
% %   \state{Ohio}
%   \country{Singapore}
% }

%% Abstract
%% Note: \begin{abstract}...\end{abstract} environment must come
%% before \maketitle command
\begin{abstract}

Graph Neural Networks (GNNs) are effective approaches for representing and analyzing many real-world graph-structured problems such as knowledge graph analysis, social networks recommendation, vaccine development and software analysis. Just like other types of neural networks such as feed-forward neural networks (FNNs) and convolutional neural networks (CNNs), the neural networks' robustness against attacks, e.g., backdoor attacks, remains an important concern.  To address this problem, property inference techniques have been proposed for FNNs and CNNs. These techniques, however, are not applicable to GNNs since they assume fixed-size inputs, as opposed to input graphs of varying structures allowed by GNNs.  

In this paper, we propose \tool{}, the first automatic property inference technique for GNNs. To tackle the challenge of varying input structures in GNNs, \tool{} first identifies a set of representative influential structures that contribute significantly towards the prediction of a GNN. 
Using these structures, \tool{} converts each pair of an influential structure and the GNN to their equivalent FNN and then leverages existing property inference techniques to effectively capture properties of the GNN that are specific to the influential structures. \tool{} then generalizes the captured properties to any input graphs that contain the influential structures. Finally, \tool{} improves the correctness of the inferred properties by building a model that estimates the deviation of GNN output from the inferred properties given full input graphs. The learned model helps \tool{} extend the inferred properties with constraints to the input and output of the GNN, obtaining stronger properties that hold on full input graphs.

Our experiments show that \tool{} is effective in inferring likely properties of popular real-world GNNs, and more importantly, these inferred properties help effectively defend against GNNs' backdoor attacks. In particular, out of the 13 ground truth properties, \tool{} re-discovered 8 correct properties and discovered likely correct properties that approximate the remaining 5 ground truth properties. Furthermore, on popular real-world GNNs, \tool{} inferred properties that can be used to effectively defend against the state-of-the-art backdoor attack technique on GNNs, namely UGBA. Experiments show that \tool{}'s defense success rate is up to 30 times better than existing defense baselines.

%\tool{} infer properties for specific \emph structures, then generalize these properties for input graphs that contain these structures.
%\tool{} obtains this set of structures by identifying the most 

%we first formally prove that given an arbitrary input graph and a GNN, there exists an equivalent FNN. Based on this theoretical result, 
\end{abstract}

  %% \keywords are mandatory in final camera-ready submission

\maketitle

\section{Introduction}
Deep neural networks (DNNs) are widely recognized as state-of-the-art in tackling a wide range of problems such as natural language processing~\cite{young2018recent, otter2020survey}, computer vision~\cite{voulodimos2018deep, liu2017survey}, and software engineering~\cite{yang2015deep, hu2018deep, pradel2021neural}. Two primary types of DNNs are (i) networks that take fixed-size input, i.e., inputs that have predefined sizes, such as feed-forward neural networks (FNNs) and convolutional neural networks (CNNs)~\cite{AlexNet, ResNetv1, ResNetv2} and (ii) networks that take inputs with varying sizes and structures, such as graph neural networks (GNNs)~\cite{Gilmer2017, Zhou2020, scarselli2008graph, Kipf2016}.

GNNs are a powerful tool for modeling complex data structures.
Problems including detecting fake users~\cite{sun2020TrustGCN}, fake news~\cite{MICHAIL2022FakenewsGCN}, and performing recommendations~\cite{ying2018graph, tan2020learning} on social media 
to mimicking classic combinatorial optimization~\cite{ijcai2021-595, Velickovic2019} and detecting faults in software~\cite{zhou2019devign, allamanis2018learning, Nguyen2022} can be straightforwardly modeled using graphs.
In these representations, a graph can have varying numbers of nodes and edges, for example, a person in social networks can have varying numbers of friends, or a molecule can have varying numbers of atoms, etc~\cite{zhou2022gnntaxnomy}. While CNNs and FNNs are not designed for such varying data structures, GNNs can naturally handle these varying structures using the message-passing mechanism~\cite{Gilmer2017}. However, the complexity of the message-passing mechanism renders it difficult to debug GNNs, e.g., assessing the robustness of GNNs against adversarial attacks~\cite{zugner2018adversarial, ma2020towards, zhu2019robust, ijcai2019-669, dai2018adversarial} or detecting backdoor behaviors~\cite{dai2023unnoticeable, Zhang2021SBA, xi2021graphtrojanattack}. To tackle GNNs debugging, existing explainable AI techniques provide heatmaps to help human experts \emph{manually} debug GNNs~\cite{ying2019gnnexplainer, luo2020parameterized, graphlime}. To the best of our knowledge, there does not exist any technique that supports automated debugging of GNNs.

There, however, exists recent works on automated debugging of CNNs and FNNs by inferring formal properties to interpret and assess deep neural network performance~\cite{Gopinath2019, Seshia2018, geng2023towards}. These techniques take as input a DNN along with a set of \emph{fixed-size} and \emph{fixed-structure} input data, and then infer a set of formal properties that are logical constraints that capture the relationships between the input and output of the DNN. Despite being powerful, these techniques cannot handle inputs with varying sizes and structures such as graphs wherein the number of nodes and edges, and the connections between them can dynamically change. It is thus significantly challenging to adopt these techniques for automated debugging of GNNs.
%GNNs take as input a graph where each node and each edge are associated with vectors representing their features and output a graph with updated node and edge feature vectors.
%The input graph to GNNs can have varying structures, e.g., the number of nodes and edges, and the connections between them can change. The structural change also leads to changes in feature vectors.
% Due to the complex and dynamic input graph structures handled by GNNs, existing DNN analysis techniques that work for FNNs or CNNs are not directly applicable to GNNs.  

In this work, we introduce \tool{}, the first automatic property inference technique for GNNs to facilitate automated debugging of GNNs. 
\tool{} tackle the challenge of dynamic input graph structures in GNN in three steps, each of which is a novel contribution. First, we select a set of representative input structures to analyze. Second, we infer GNN's properties on each of the representative structures. Finally, we generalize these properties to varying structures that contain the representative structures. To achieve these, \tool{} first mines the \emph{most frequent influential} structures, which appear frequently in the dataset and contribute significantly towards GNN's prediction. These structures are more likely to capture GNN's behavior as shown in prior works~\cite{Ying2019, ma2022influentialattackgraph}, and are thus considered representative input structures.
%\dat{We formally prove that given a GNN model $\mathcal{M}$ with \dat{any chosen} input graph $S$ is recursively \dat{\textit{reducible} to an equivalent} feed-forward neural (FNN) model $\mathcal{M}_F$ and describe give the algorithm to perform this transformation. }
Next, for each influential structure $S$ as input to a GNN model $\mathcal{M}$, \tool{} converts the pair $\langle \mathcal{M}, S\rangle$ into an equivalent FNN model $\mathcal{M}_F$. We formally prove that our conversion algorithm is both sound, i.e., $\mathcal{M}_F$ is indeed equivalent to $\langle \mathcal{M}, S\rangle$, and complete, i.e., given any $\langle \mathcal{M}, S\rangle$, there always exists an equivalent $\mathcal{M}_F$. This enables us to leverage an existing property inference technique for FNNs, namely PROPHECY \cite{Gopinath2019}, to infer a set of properties $\{S_P\}$. These inferred properties $\{S_P\}$ are then considered the properties of the original GNN $\mathcal{M}$ on the given structure $S$. 

%Each of $S_P$ has the following form: $\sigma_{S\simeq} \land \sigma_{inps} \Rightarrow P$ where $\sigma_{S\simeq}$, $\sigma_{inps}$ and $P$ are the structural condition, the input feature condition, and the output condition respectively.
%The structural condition $\sigma_{S\simeq}$ specifies that the property $S_P$ applies only for input graphs that are isomorphic graphs of $S$.
%The input feature condition $\sigma_{inps}$ specifies a convex region over the node and edge features.
%Finally, the output condition $P$ specifies the condition over the prediction value of GNN. $P$ either specifies the output being a chosen class in classification problems or a linear relation between the output and the input features for regression problems. 
%The structural and input feature conditions specify constraints/preconditions on the input for the output condition to hold. Hence, they are also referred to as constraints or predicates.
%These inferred properties are verifiable by a verification tool for FNNs and CNNs, namely Marabou~\cite{Katz2019}.

Note that the inferred properties $\{S_P\}$ are still specific to the frequent influential structures: each property $S_P$ imposes an input condition that the input graph must have the \emph{exact} structure as $S$.
\tool{} then generalizes these properties to any input graph that contains the influential structures by relaxing each $S_P$ into a likely property $S_{LP}$, in which the imposed input condition is that the input graph must \emph{contain} the structure $S$.
%The relaxed likely property $S_{LP}$ has the following form $\sigma_{S\preceq} \land \sigma_{inps} \Rightarrow P$.
%We do this by relaxing the structure condition $\sigma_{S\simeq}$ into $\sigma_{S\preceq}$ for each structure-specific property $S_P$. 
%The relaxed structure condition $\sigma_{S\preceq}$ specifies that the input graph has to \emph{contain} the influential structure $S$, i.e., the input graph is \emph{subgraph isomorphic} with $S$. 
When the imposed input condition is relaxed, the likely property $S_{LP}$ might lose precision. \tool{} improves the precision of these likely properties by training a model (either a decision tree or linear regression) that estimates the deviation of the GNN output from the inferred properties on a given full input graph. The predictions of the model help \tool{} obtain more precise properties that hold on the full input graph.

We evaluate \tool{}'s correctness and applicability on both synthetic and popular real-world datasets.
To assess \tool{}'s correctness, we construct a synthetic benchmark consisting of reference GNN and corresponding ground truth properties: the constructed benchmark consists of reference and trained GNNs to mimic classic breadth-first search (BFS) and depth-first search (DFS) and Bellman-Ford (B-F) algorithms. Manual assessments show that out of 13 ground-truth properties, \tool{} re-discovered 8 correct properties and identified highly confident properties that approximate the remaining ground-truth properties.
As an application of \tool{}, we use its inferred properties to detect and defend against the state-of-the-art backdoor attack technique on GNNs, namely UGBA~\cite{dai2023unnoticeable}. Experiments show that \tool{}'s inferred properties can be used to improve the defense against the state-of-the-art backdoor attack method UGBA~\cite{dai2023unnoticeable} by up to 30 times in comparison with existing baseline defense methods~\cite{dai2023unnoticeable}. In summary, we make the following contributions:
\begin{itemize}
    \item We propose \tool{} which automatically infers properties of GNNs by analyzing and generalizing GNNs' behaviors on representative structures.
    \item We provide an algorithm that converts a GNN with a given input structure to an equivalent FNN. We formally prove that the algorithm is both sound and complete, enabling us to use existing DNN analysis tools to infer properties of the GNN.
    \item We construct a set of benchmarks containing GNNs on classic traversal and shortest path algorithms to evaluate \tool{} and show that \tool{} can infer correct properties.
    \item Finally, we show that \tool{} can be used to improve the backdoor defense against state-of-the-art backdoor attack techniques on graph neural networks by up to 30 times in comparison with existing techniques, namely homophily-based pruning and isolation~\cite{dai2023unnoticeable}.
\end{itemize}

%\tvn{I usually do not include this kind of roadmap. People can quickly skim through the paper to figure out what these sections are. I would just remove it. Use the space for other more insightful discussions} \dat{I have commented it out.}
%The organization of the paper is as follows. Section~\ref{sec:background_gnn} describes the background on FNN and GNNs. Section~\ref{sec:overview} gives an overview and running example of \tool{}. Section~\ref{sec:model} depicts the detail of the \tool{} approach. Section~\ref{sec:eval} describes our experiments and results. Section~\ref{sec:related} describes the related works and Section~\ref{sec:conclusion} concludes our findings and future works.

\section{Background}\label{sec:background_gnn}
The key idea of \tool{} is based on transforming a GNN into an FNN. We now give a brief description of FNN and GNN.

\noindent\textbf{Feed-forward Neural Networks~\cite{Singh2019}}.
Definition~\ref{def:ffnn_layer} and Definition~\ref{def:ffnn} below give a formal definition of Feed-forward Neural networks.

\begin{definition} \label{def:ffnn_layer} A feed-forward layer is a function between $m$ input variables $x_{1}, .., x_{m}$ and $n$ output variables $y_{1}, .., y_{n}$. This function is one of the following:
\begin{enumerate}
    \item an affine transformation: $(y_1, \ldots y_n)^\top = \mathbf{A} (x_1, \ldots x_m)^\top + \mathbf{b}$.
    \item the $\mbox{ReLU}$ activation function $(y_1, \ldots y_n)^\top= \left(\max(0, x_1) \ldots \max(0, x_m)\right)^\top$. Note that for $\mbox{ReLU}$ the number of input and output variables has to be the same (i.e., $m = n$).
    \item a max pooling function, $(y_1, \ldots, y_n)^\top = (max_{x \in P_1}(x), \max_{x \in P_2}(x), \ldots, \max_{x \in P_n}(x))^\top$ where $P_{1}, \ldots P_n$ are predefined subsets of the input variable set $\{x_1, \ldots, x_m\}$. 
\end{enumerate}
\end{definition}

\begin{definition} \label{def:ffnn} (FNN) An $l$-layered FNN is a composition of $l$ feed-forward layers.
\end{definition}

Intuitively, an FNN is a composition of specific operations applied to the input. The input of FNN is a vector $\mathbf{x} = [x_1, \ldots, x_m]^\top \in \mathbb{R}^m$ where $m$ is the chosen dimension for input. Given this input, the applied operations are either linear affine transformations or non-linear activation functions. The output of FNN is the output vector $\mathbf{y} = [y_1, \ldots, y_n]^\top \in \mathbb{R}^{n}$, where $n$ is the chosen dimension for output (i.e., number of classes in a classification problem).

\noindent\textbf{Graph Neural Network~\cite{Gilmer2017}}. Graph Neural Networks (GNNs), on the other hand, are a more recent development in the field of neural networks that focus on graph-structured data. They take as input a graph $G$, where each node and edge can be associated with their respective feature vector and make predictions on the graph, nodes or edges' properties. Similarly to FNN, GNN (see Definition~\ref{def:gnn}) is a composition of message passing layers defined in Definition~\ref{def:message_passing_layer}.

\begin{definition} \label{def:message_passing_layer} (Message Passing Layer) A graph neural network layer is a function that takes as input a graph $G=(V, E)$, where $V$ is the set of nodes and $E$ is the set of edges, along with associated node feature $\mathbf{x}_i$ for each node $i$ and edge feature $\mathbf{e}_{ji}$ for each edge $(j, i) \in E$, and return the updated node features $\mathbf{x}'_i$ for each node $i \in V$. The update is based on the node's previous state and the aggregated states of its neighboring vertices and edges. Formally, the update rule for a node $i$ can be defined as:
\[ \textbf{x}_i' = f_{upd} \left( \textbf{x}_i, f_{agg} \left( \left\{ f_{msg}( \textbf{x}_{j}, \textbf{x}_i, \textbf{e}_{ji}) \forall j | (j, i) \in E  \right\} \right) \right) \]
where $f_{agg}$ is an aggregation function, and $f_{upd}$ is an update function and $f_{msg}$ is a message passing function. $f_{upd}$ and $f_{msg}$ are FNNs according to Definition~\ref{def:ffnn} and $f_{agg}$ is either a linear combination or max aggregation over a set of vectors.
\end{definition}

\begin{definition} \label{def:gnn}
A Graph Neural Network (GNN) is a neural network that operates on the graph domain and consists of a sequence of graph neural network layers. Each layer updates the vertices' states based on the graph structure and vertex and edge features. An $l$-layered GNN is a composition of $l$ such layers, with the output of each layer serving as the input to the next layer.
\end{definition}

A GNN takes as input a graph $G = (V, E)$ with associated node features $\mathbf{x}_i$ for each node $i \in V$, and edge features $\mathbf{e}_{ji}$ for each edge $(j, i) \in E$ where $V$ is the node set and $E$ is the set of edges, where each node and edge is associated with their own feature vectors. The GNN iteratively computes new features for nodes via message-passing layers. Each message passing layer first calculates the message $m_{j, i}$ for each edge $(j, i) \in E$ by making use of the message passing function $f_{msg}$. Following that, the aggregation function $f_{agg}$ comprises the incoming messages (i.e., vectors) for each individual node $i$ into an aggregated message (a single vector). Finally, this aggregated message is transformed by another update function $f_{upd}$, resulting in the updated node features.

\begin{figure}
\centering
\small
    \includegraphics[width=\linewidth]{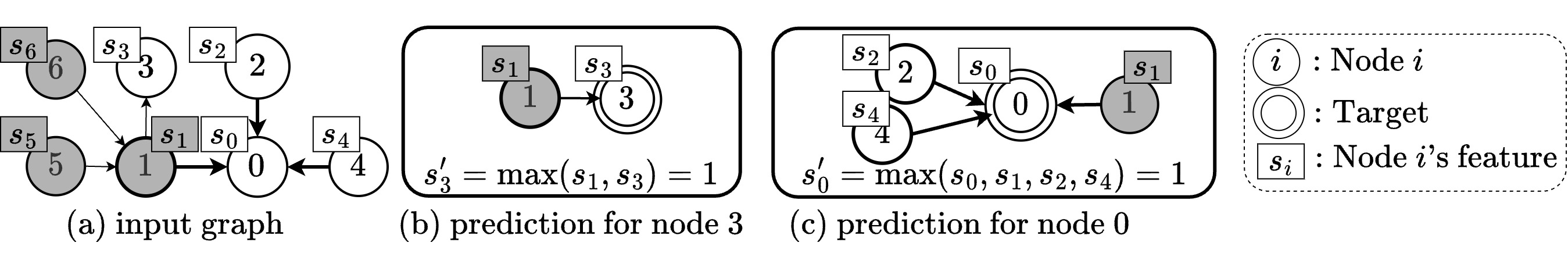}
    \caption{Message passing process of BFS. While the GNN makes predictions in parallel for all nodes, we separate nodes $0$ and $3$'s predictions for illustration shown in (b) and (c).}
		\label{fig:motiv_bfs}
\end{figure}

As an example, we take a GNN simulating Breadth-first search (BFS) algorithm in (Fig ~\ref{fig:motiv_bfs}), following~\cite{Velickovic2019}. This GNN takes as input a graph, where each node $i$ is associated with a single feature $s_i \in \{0, 1\}$ denoting whether the node has been visited ($1$) or not ($0$), and predicts these nodes' visiting state in the next step. Let the message and update functions be an identity function (i.e., $f(x) = x$), the aggregation is chosen as the max aggregation.
With this combination, each individual node's update features will be calculated as $s'_i = \max\{s_j | (j, i) \in E\}$, meaning that each node's new features is the maximum among its neighbors features.
To see how this GNN works, let us take as input a graph with 7 nodes in Fig ~\ref{fig:motiv_bfs}. nodes $3$, receiving incoming edges from node $1$ and node $0$, receiving incoming edges from nodes $1, 2, 4$,  are predicted as \emph{visited} (Fig.~\ref{fig:motiv_bfs}) since node 1, among their neighbor is visited. In practice, this message-passing mechanism allows GNN to handle arbitrary input graphs, of which sizes and structures directly affect the computation. Due to the complexity of the message-passing mechanism, there does not exist formal techniques and tools for analyzing GNNs.

%From this point of view, FNN can be seen as a reduced GNN handling a single node with varying features and without any edge.

Our \tool{} approach aims to reduce a GNN to an FNN, and take advantage of existing FNN analyses to infer properties for GNNs.
Note that for illustration, the above example gives a very simple message-passing procedure that is just identity functions. Our approach \tool{} indeed supports even more complicated message-passing procedures, e.g., $f_{\text{msg}}$ and $f_{\text{upd}}$ can be FNNs themselves and can operate on a heterogeneous graph similar to those in the literature~\cite{Schlichtkrull2017, Zhao2021}.  In Section~\ref{sec:overview}, we give an overview of \tool{} and how it works on the BFS example in Fig.~\ref{fig:motiv_bfs}.

\section{Overview and Motivating Example}\label{sec:overview}
\begin{figure}
    \centering
    \small
    \includegraphics[width=0.85\linewidth]{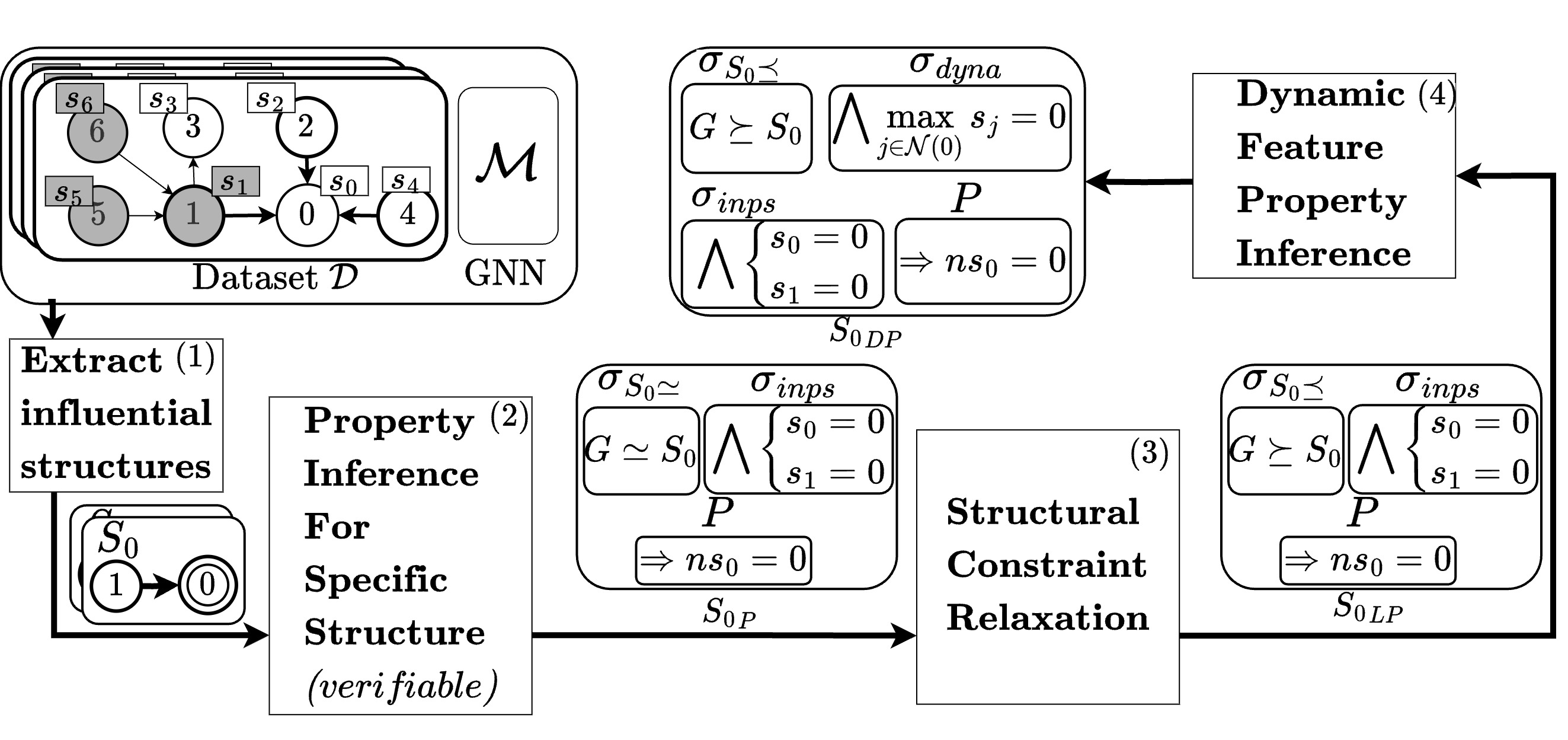}
    \caption{\tool{} Overview.}
    \label{fig:overview}
\end{figure}

%\tvn{See if you can move overview before Background. Overview is very short now so you can combine it with Background, which appears after overview.}
Fig.~\ref{fig:overview} presents an overview of \tool{}, which takes as inputs a trained GNN model $\mathcal{M}$ and a dataset $\mathcal{D}$, and outputs likely GNN properties and structure-specific properties. \tool{} comprises four stages: (i) extracting influential substructures, (ii) inferring properties for specific structures, (iii) relaxing structure constraints, and (iv) inferring dynamic feature properties. Below, we explain these steps and also how it works on the BFS example presented in Fig.~\ref{fig:motiv_bfs}.

\begin{figure}
	\centering
 \small
	\includegraphics[width=0.7\linewidth]{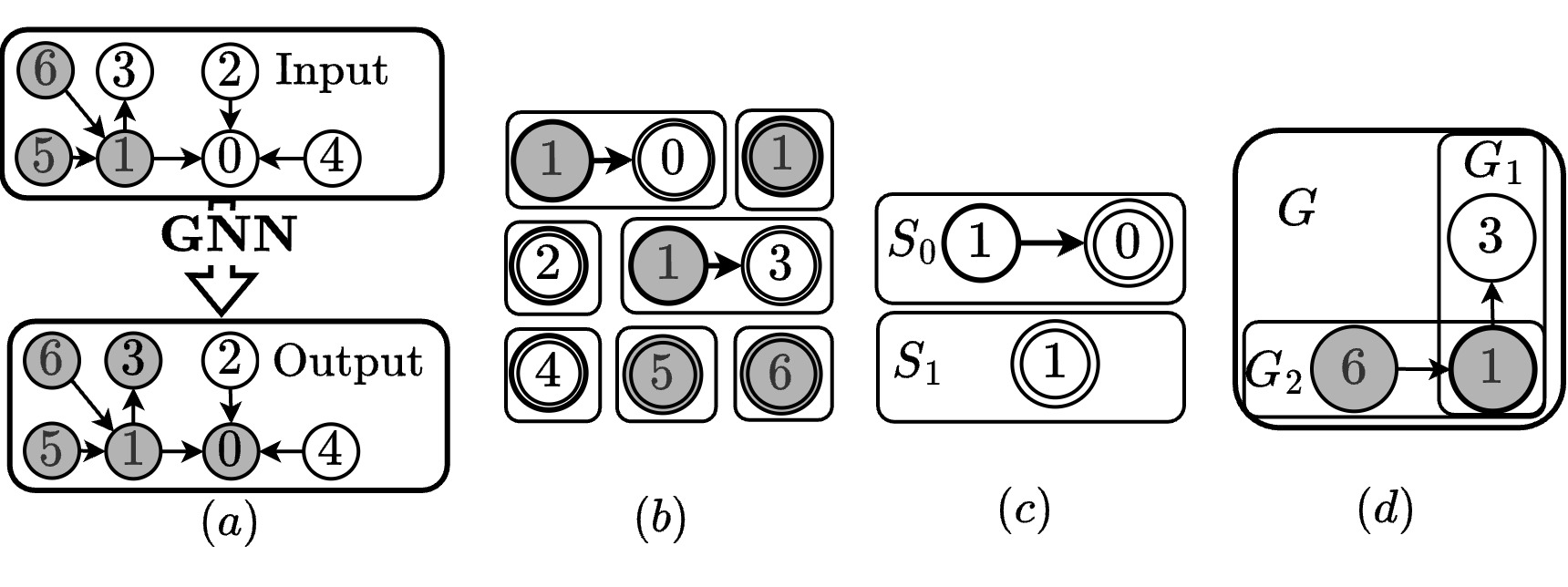}	\caption{(a) a GNN step, visited nodes are grey-colored (b) Example of influential structure for each node for the GNN step, where each influential structure contributes significantly to the target node denoted in the double circles. (c) the most frequented influential structures and (d) graph matching of between $S_0$ and $G$.}
	\label{fig:example_infl_structure}
\end{figure}

\noindent\textbf{Extract influential structures}. This step analyzes the model $\mathcal{M}$ and dataset $\mathcal{D}$ to compute influential structures, which are subgraphs that frequently appear in the dataset and significantly impact GNN predictions. \tool{} identifies influential structures for individual nodes using GNNExplainer~\cite{ying2019gnnexplainer} and gSpan~\cite{yan2002gspan}. For example, in the BFS GNN prediction of Fig.~\ref{fig:example_infl_structure}a, \tool{} first detects influential structures for each node. In this case, the influential structure for node $0$ is a subgraph containing node $1$ and node $0$ with an edge from node $1$ to $0$, because node $0$ is predicted as visited due to its connection with the visited node $1$. For node $1$, its influential structure is just itself, since it would remain to be visited regardless of its neighbors' states. The result of applying this process to all nodes is illustrated in Fig.~\ref{fig:example_infl_structure}b. Then, from these influential structures of all nodes, \tool{} chooses the two most frequently appeared structures $S_0$ and $S_1$ in Fig.~\ref{fig:example_infl_structure}c.

\noindent\textbf{Property inference for specific structure}. For each influential structure $S$, \tool{} infers a set of structure-specific properties $\{S_P\}$, each in the form of $\sigma_{S\simeq} \land \sigma_{inps} \Rightarrow P$. $\sigma_{S\simeq}$ is a structure condition requiring the input graph to be isomorphic with $S$, $\sigma_{inps}$ is an input feature condition, and $P$ is an output condition on prediction values. \tool{} uses Neo4J~\cite{neo4j} to collect all graph matching the structure condition $\sigma_{S\simeq}$ from the dataset $\mathcal{D}$, transforms the GNN model $\mathcal{M}$ and structure $S$ into an equivalent model $\mathcal{M}_F$, and employs PROPHECY~\cite{Gopinath2019} to identify $\sigma_{inps}$ and $P$.

For the BFS example, \tool{} derives properties for substructure $S_0$ for node $0$ like $\sigma_{S_0\simeq} \land s_0 = 0 \land s_1 = 0 \Rightarrow s'_0 = 0$ and $\sigma_{S_0\simeq} \land s_0 = 1\Rightarrow s'_0 = 1$. These indicate that for input graphs isomorphic with $S_0$ (containing edge $1 \rightarrow 0$), the GNN predicts node 0 is unvisited if both node 1 and itself are unvisited, and visited if node 1 is visited. For substructure $S_1$, \tool{} obtains a property stating that for input graphs isomorphic with $S_1$, node 1's predicted status equals its current status.

The found properties apply only to graphs isomorphic with influential substructures, enabling \tool{} to use automatic verification like Marabou~\cite{Katz2019}. However, these structure-specific properties are limited (e.g., only applicable to the graphs with this exact structure).

\noindent\textbf{Structure Constraint Relaxation}. \tool{} generalizes each structure-specific property to a \emph{subgraph isomorphic} property, covering a broader set of input graphs. We obtain properties like $\sigma_{S\preceq} \land \sigma_{inps} \Rightarrow P$, where $\sigma_{S\preceq}$ denotes subgraph isomorphism condition (i.e., input graph has to contain structure $S$). 
In the BFS example, \tool{} derives $\sigma_{S_0\preceq} \land s_i = 0 \land s_j = 0 \Rightarrow s'_i = 0$, which means that any input graph that \textit{contain} this structure $S_0$, with both node $i$ and $j$ (neighbor of node $i$) being unvisited, then the new state of node $i$ will also be \textit{unvisited}.
These isomorphic properties are more general and useful but, as they cover dynamic structures, GNN-to-FNN conversion and verification using existing DNN tools are not possible, making them likely properties.

\noindent\textbf{Dynamic Property Inference}.
%\tvn{this is mouthful, can you simplify it to just dynamic feature or dynamic property inference?}\dat{I reduced it to dynamic property inference}
To enhance the precision of likely properties, \tool{} augments them with dynamic feature properties, capturing output postcondition changes relative to an aggregated feature set of the GNN. To do this, \tool{} computes feature formulae from the trained GNN, collects feature values for input graphs in dataset $\mathcal{D}$, and trains a decision tree or linear regression on them. For classification problems, \tool{} predicts whether condition $P$ holds on the full graph and constructs dynamic feature predicate $\sigma_{dyna}$ using tree paths leading to $P=True$. The dynamic property is $\sigma_{S\preceq} \land \sigma_{inps} \land \sigma_{dyna} \Rightarrow P$. For regression problems, \tool{} trains a linear regression model to capture deviation between condition $P$ and GNN's actual output, obtaining new output condition $P_{dyna}$ and dynamic property $\sigma_{S\preceq} \land \sigma_{inps} \Rightarrow P_{dyna}$.

Using the BFS example, \tool{} extracts the dynamic feature $\max_{j \in \mathcal{N}(i)} s_j$ for nodes in structure $S$. This feature incorporates surrounding node information. \tool{} collects aggregated feature values and builds a decision tree to predict property applicability. The logical formula is combined with the structure-property to obtain the full property, ensuring accurate predictions. If the property already matches all instances in the dataset, no dynamic condition is added.

% \tvn{provide a summary statement, at the end, what results does \tool{} give?  what does it mean and why it is good?}
Finally, the results of \tool{} are likely properties in either the form of $\sigma_{S\preceq} \land \sigma_{inps} \land \sigma_{dyna}\Rightarrow P$ for classification problems or $\sigma_{S\preceq} \land \sigma_{inps} \Rightarrow P_{dyna}$ for regression problems. These properties show that if the input graph containing the structure $S$, having input features matching $\sigma_{inps}$ and $\sigma_{dyna}$, then the output property ($P$ or $P_{dyna}$) is implied. These properties show specific behaviors of the GNN on a class of graphs and can be used for precise analysis of GNN, one example is using these properties to find backdoor behaviors and prune the backdoored graph (See Section~\ref{sec:backdoor})

\section{The \tool{} Approach}\label{sec:model}

We now present in more details of our approach \tool{} illustrated in Fig.~\ref{fig:overview_algo}. \tool{} takes a trained GNN model $\mathcal{M}$ and dataset $\mathcal{D}$ as inputs
% \tvn{this paragraph does not mention what $\mathcal{D}$ is used for}
. First, we extract influential structures $\{S\}$ given the model $\mathcal{M}$ and dataset $\mathcal{D}$. Followed by that, we convert each $\langle \mathcal {M}, S \rangle$ to an equivalent FNN $\mathcal M_F$, collecting $\mathcal M_F$'s execution trace given dataset $\mathcal{D}$ and inferring structure-specific properties $\{S_P\}$ given the obtained execution traces. Finally both $\mathcal{M}$ and $\mathcal{D}$ are used to generalize each structure-specific property $\{S_P\}$. Note that \tool{} works on any $\mathcal{D}$ that are graphs, and we show in Section~\ref{sec:eval} that \tool{} works well on existing popular dataset and GNN models. Next subsections explain each step of \tool{} in details.

%\tvn{A  concern a PLDI reviewer might raise is that everything here depends on this dataset, how "good"/"big" does it have to be?  you should say something about it here, e.g., we show in Eval that the dataset doesn't have too be too big or too good to obtain good results}

%\tool{}'s novelty lies in three steps. At the first step, \tool{} uses influential structures as representative structures for better generalizability and efficiency. At the second step, we introduce algorithm to convert any pair $\langle \mathcal M, \mathcal D \rangle$ to an equivalent FNN $\mathcal{M}_F$. We formally prove that the algorithm is both sound and complete (See Section~\ref{sec:struct_specific_analysis}). These steps enable us to leverage existing property inference tools to infer structure-specific properties. Finally, to generalize structure-specific properties in the dynamic property inference step, we propose using features extracted automatically from the first message passing layer of GNN, this allows \tool{} to capture the difference that can occur when relaxing the transiting from analyzing specific structures to all the graphs that can contain these structures.}

%\tvn{Say something else here -- what do using these fancy things allow you to do ?}\dat{I have updated the paragraph accordingly.}

\DontPrintSemicolon
\begin{algorithm}[t]
\small
	\SetAlgoLined
 
	\SetKwInOut{Input}{Input}\SetKwInOut{Output}{Output}
	\Input{A trained GNN model $\mathcal{M}$, training dataset $\mathcal{D}$}
	\Output{A set of structure-specific properties $SP$, and a set of dynamic feature properties $DP$}

	$IS \gets \texttt{ExtractStructures}(\mathcal{M}$, $\mathcal{D})$\;
	$SP \gets\{\}$, $DP \gets \{\}$\;
	\ForEach{\text{structure} $S \in IS$}{

		$\sigma_{S\simeq}\gets \texttt{ConstructGraphIsomorphicPredicate}(S)$\; 
		$\sigma_{S\preceq} \gets \texttt{ConstructSubgraphIsomorphicPredicate}(S)$\;
            $\{S_P\} \gets \texttt{InferStructSpecificProps}(\sigma_{S\simeq}, \mathcal{M}, \mathcal{D})$\;
    \ForEach{$S_P \in \{S_P\}$}{
			$SP$.add($S_P$)\;
                $S_{LP} \gets \texttt{RelaxConstraint}(S_P, \sigma_{S\preceq})$\;
                $S_{DP} \gets \texttt{DynamicAnalysis}(S_{LP}, \mathcal{M}, \mathcal{D}$)\;			
                $DP$.add($S_{DP}$)\;
		}
	}
        \Return $SP, DP$ \;
	\caption{The \tool{} approach}
	\label{fig:overview_algo}
\end{algorithm}

\subsection{Extracting Influential Substructures} \label{subsec:infl_extract}
The use of \emph{most frequent influential structures} empowers \tool{} to more efficiently and effectively analyze a GNN. Our intuition is that:
(1) The frequent influential structures are smaller than full graphs, rendering them more efficient to analyze. (2) They contribute significantly to the GNN's prediction, and hence, they help reveal common behaviors of the GNN. Towards this end, \tool{} uses GNNExplainer~\cite{Ying2019} to identify the influential structures and Subdue~\cite{Ketkar2005} to find the most frequent structures among the set of all influential structures.

%We choose influential structures to analyze due to two characteristics besides having fixed size (which is the requirement to convert a GNN into FNN): 
% in comparison with other parts of the input graph.
%While using influential structures helps to identify important subgraphs, there can still be a large amount of distinct influential structures, and analyzing all of them can be computationally expensive. 
%We balance the coverage and the computation cost by analyzing only the most frequented influential structures. 
%These structures have high coverage (appears frequently in $\mathcal{D}$) which leads to improved generalizability.
%Furthermore, they are lower in numbers, leading to reduced computation cost in analysis.
% Since these frequented structures have high coverage (i.e., the number of graphs in the dataset that contain these structures) and are lower in numbers, analyzing GNNs' behavior on these structures leads to improved generalizability and reduced computation cost.

While the GNN makes predictions for all nodes of the input graph in parallel, each node's prediction can be influenced by different parts of the input graph as shown in Fig.~\ref{fig:example_infl_structure} and Section.~\ref{sec:overview}.
To precisely capture the influential structures of the GNN, we identify the influential structure for each node prediction individually. The chosen node is called the \emph{target} node.
This is done by applying GNNExplainer~\cite{Ying2019} on the target node prediction. The GNNExplainer takes as input a graph, the model $\mathcal{M}$, and the target prediction and outputs important scores for each node and edge in the input graph towards the prediction. We retrieve the influential structure from the important score by keeping the nodes and edges having sufficient important scores (above the threshold of $0.8$ in our implementation). Retrieving each node's influential structure would give us a set of \emph{influential structures} $IS$ in the dataset $\mathcal{D}$ with respect to the model $\mathcal{M}$.
On this set $IS$, we apply gSpan~\cite{yan2002gspan} to obtain a small set of \emph{frequent influential structures}.

\paragraph{Example} Taking the example in Fig. ~\ref{fig:example_infl_structure}, each node $0, 1, 2, 3, 4, 5, 6$ has a distinct influential structure. 
Each of these influential structures is associated with a target node and contains information on the set of nodes, edges, node types, and edge types (without node or edge features). 
These structures are smaller than the input graph (e.g., node $0$'s influential structure consists of only two nodes and one edge in comparison with $7$ nodes and $6$ edges of the original input graph).
Performing frequented subgraph mining would give us 2 most frequented influential structures: (1) The structure has a single node and (2) the structure has two nodes (e.g., node $0$'s influential structure).
This set of frequent influential structures is used in a later step as structure predicates described in Section~\ref{subsec:struct_predicate}, which are used to govern the constraint over which graph structures to be analyzed.

\subsection{Structure Predicates} \label{subsec:struct_predicate}
Structure predicates are used to describe the constraints over which graph structures that \tool{} focuses on analyzing. Below, we provide a formal definition of structure predicates.

Recall that a GNN takes as input a graph $G = (V_G, E_G)$. Each influential structure $S = \langle V_S, E_S \rangle $ is a subgraph of $G$, and is influential towards a single prediction.
 Since \tool{} performs properties inference for specific structures, and subsequently generalizes the input for the structures that contain these specific structures, we use isomorphism and subgraph isomorphism as structure predicates.

Intuitively, given a graph $G'$, an isomorphism predicate on a graph $G$ checks whether $G'$ and $G$ has identical structure. We define graph isomorphic as $\simeq$ and the $G'$-isomorphic predicate as $\sigma_{G'\simeq}$ as follows:
\begin{equation}
    \sigma_{G'\simeq}(G) := G' \simeq G := \exists h: V_G \to V_G' \text{ s.t. } \begin{array}{l} \forall i \in V_G, i \in V_G' \\ \forall i, j \in E_G, (h(i), h(j)) \in E_{G'} \text{ and }\\ \forall (i, j) \not \in E_G, (h(i), h(j)) \not \in E_{G'} \end{array}
\end{equation}
This indicates that there must exist a bijective mapping between the nodes of $G$ and $G'$ such that if a node or an edge exists in $G$, it must have the corresponding node or edge in $G'$ and vice versa. This isomorphic condition is strict and requires that there exists no redundant edge in $G$ (i.e., $G$ and $G'$ have to be identical in term of structure). We use this as a structure predicate for the structure-specific properties.

Additionally, if a graph $G$ contains a subgraph that is isomorphic to $G'$, this would result in subgraph-isomorphic condition $\preceq$. Specifically, let us also define the subgraph isomorphic operation as $\preceq$ and the $G'$-subgraph isomorphic predicate $\sigma_{G'\preceq}$ as follows:
\begin{equation}
    \sigma_{G'\preceq}(G) := G' \preceq G := \exists h: V_G' \to V_G \text{ s.t. } \begin{array}{l} \forall i \in V_{G'}: h(i) \in V_G \\ \forall i, j \in E_{G'}, (h(i), h(j)) \in E_{G}\end{array}.   
\end{equation}
This definition specifies that there exists an injective function $h$ from the node set of $G'$ to the node set of $G$ such that every edge $G'$ corresponds to an edge in $G$, thus ensuring that the mapped vertices from $G$ form a subgraph isomorphic to $G'$. Intuitively, this can be understood as a more relaxed version of graph isomorphic condition. 

Structure predicates are then formally defined as:
\begin{equation}
    \sigma_{struct} \coloneqq \left\{ \begin{array}{ll}
    \sigma_{S\simeq} & \text{When using graph-isomorphic constraint}\\
    \sigma_{S\preceq} & \text{When using subgraph-isomorphic constraint}
    \end{array} \right.
\end{equation}

%Given a specific property with a structure predicate requiring that $G$ having identical structure as $G'$, the last step of \tool{} is finding under which feature condition, the graph-isomorphic condition 

\paragraph{Example} Consider the structure $S_0$ illustrated in Fig. ~\ref{fig:example_infl_structure}. We define $S_0 = \langle \{0, 1\}, \{(1, 0)\}\rangle $, having 2 nodes 0, 1 and one edge from 1 to 0 and $S_0$ is influential towards the prediction of target node 0. $\sigma_{S_0\simeq}(G)$ and $\sigma_{S_0\preceq}(G)$. In this example, graph $G$ is not isomorphic towards $S_0$. But $G$ contains two subgraphs $G_1$ and $G_2$ that is isomorphic towards $S_0$. Hence, $\sigma_{S_0\simeq}(G_1) = True$ and $\sigma_{G_0\simeq}(S_2) = True$. Moreover, $G_1$, $G_2$, and $G$ are all subgraph-isomorhic towards $S_0$, thus, $\sigma_{S_0\preceq}(G_1) = \sigma_{S_0\preceq}(G_2) = \sigma_{S_0\preceq}(G) = True$.

\subsection{Property Inference for Specific Structure}\label{sec:struct_specific_analysis}

This step takes the GNN $\mathcal{M}$, the dataset $\mathcal{D}$, a structure $S$ and outputs a set of structure-specific properties $\{S_P\}$. Each structure specific property $S_P$ have the form of $\sigma_{S\simeq} \wedge \sigma_{inps} \Rightarrow P$. Here, $\sigma_{S\simeq}$ specifies graph-isomorphic predicate in Section ~\ref{subsec:struct_predicate}, while $\sigma_{inps}$ is the input feature predicate, and $P$ is the output property.
The input feature predicate $\sigma_{inps}$ specifies a set of linear inequalities on the input features of each node $\mathbf{x_i}$ and each edges $\mathbf{e}_{ji}$, while the output property $P$ either specifies a chosen output class of each node for classification problems or a linear relation between the output and the input for regression problems.

To achieve the above, \tool{} first transforms the pair $\langle \mathcal{M}, S \rangle$ into a corresponding FNN $\mathcal{M}_F$. We prove that this transformation is both sound and complete (See Section~\ref{subsec:rollout}). Based on this theoretical foundation, we then use 
PROPHECY~\cite{Gopinath2019}, a property inference tool for FNN, to infer the properties of $\mathcal{M}_F$. The inferred properties are then considered to be properties of the original GNN $\mathcal{M}$.

%\dat{As the basis of this step, we formally demonstrate that a GNN in Definition~\ref{def:gnn}, given the chosen, fixed structure $S$ is reducible to an FNN in Definition~\ref{def:ffnn}.
%Based on this theoretical base, we demonstrate the transformation from a pair of GNN models and structure $(\mathcal{M}, S)$ into its equivalent FNN model $\mathcal{M}_F$. 
%Obtaining an equivalent FNN model $\mathcal{M}_F$ allowing us to use PROPHECY~\cite{Gopinath2019} to directly infer the properties of $\mathcal{M}$, given the structure (i.e., set of nodes and edges in $S$) as the partial input.}

Note that the inferred properties are specific to the input structure $S$. Since PROPHECY's inferred properties are guaranteed to be correct (verified by the Marabou tool~\cite{Katz2019} used in PROPHECY), these properties are correct on the original GNN when the input graph is isomorphic with $S$.

\subsubsection{Theory and proof of transformation from GNNs to FNNs }\label{subsec:rollout}

This section demonstrates a sketch proof that, for any given Graph Neural Network (GNN) model $\mathcal{M}$ following Definition~\ref{def:gnn}, and an input structure $S = (V_S, E_S)$, the pair $\langle \mathcal{M}, S\rangle$ is reducible to an equivalent FNN $\mathcal{M}_F$.

We start the proof by introducing two lemmas to show that (1) the composition of different neural networks is a neural network (2) and also, the layer-wise combination of different $l$-layer neural networks is a $l$-layer neural network. 
In the context of our work, we define a layer-wise combination of different $l$-layer FNNs in Definition ~\ref{def:combination_ffnn} below.

\begin{definition}\label{def:combination_ffnn} (Layer-wise combination of FFNs) Given a set of $l$-layered FNNs $\{\mathcal{M}_{F_1}, \mathcal{M}_{F_2}, .. \mathcal{M}_{F_n}\}$, each $\mathcal{M}_{F_i}$ defines a relation between its input and output variables, a layer-wise combination of $\{\mathcal{M}_{F_1}, \ldots, \mathcal{M}_{F_n}\}$ is a composition of $l$ relations $L^{(1)}, \ldots L^{(l)}$, where each relation $L^{(i)}$ is formed by combining each $i$th layer of all the models $\mathcal{M}_{F_1}, \ldots, \mathcal{M}_{F_n}$: $L^{(i)}$’s input and output variable sets are unions of the input and output variable sets of all layers forming it respectively and the relation between these input and output variable sets is defined FNNs' layers. 
\end{definition}

This composition is common in FNN, demonstrating the process in which the results of each layer can be concatenated, added in different FNN architectures to form the final results. We also show that every layer-wise combination of an FNN results in an FNN itself in Lemma~\ref{lem:combine} below.

\begin{lemma}
    \label{lem:combine}
    The layer-wise combination of $n$ $l$-layered FNNs $\mathcal{M}_{1}, \ldots, \mathcal{M}_{n}$ is a FNN.
\end{lemma}

Each layer's combination produces a new layer that computes relations depending on the combination of the input and output variable sets of the corresponding layers from the original FNNs. Furthermore, we can show that the composition of multiple FNNs also result in another FNN according to Lemma~\ref{lem:composition}.

\begin{lemma}
    The composition of $k$ Feed-forward Neural Network $\mathcal{M}_{1}, \cdots, \mathcal{M}_{k}$, where $\mathcal{M}_{i}: {R}^{n_{i}} \rightarrow \mathbb{R}^{n_{i+1}}$, is a Feed-forward Neural Network.
    \label{lem:composition}
\end{lemma}

This establishes the basis for combining different neural network layers while preserving the FNN structure. Lemma \ref{lem:composition} affirms that the composition of $k$ FNNs results in another FNN. %Since the composition of functions is a fundamental operation in FNNs, this lemma supports the notion that sequentially processing data through layers or composed networks is equivalent to using an FNN.
Next, we show that, given a fixed structure as input to a GNN, every layer of the GNN is reducible to a FNN:

\begin{lemma} \label{lem:agg_reduction} Given a graph $G$ with fixed structure (i.e., the edge set $E$ remains unchanged), and the hidden features $\mathbf{x}_i \in \mathbb{R}^D$ is produced by a feed-forward neural network in Definition~\ref{def:ffnn}, a graph neural network's message passing layer as defined in Definition~\ref{def:message_passing_layer} is reducible to a feedforward neural networks defined by Definition~\ref{def:ffnn}.
\end{lemma}

This lemma provides a critical reduction, indicating that a layer in a GNN, even when the graph structure is fixed, can be considered analogous to a layer in an FNN. Given that GNNs operate on graph structures and FNNs on flat inputs, this bridges the gap by showing that the graph-based message-passing procedure can be reformulated as FNN operations. Finally, we establish Theorem~\ref{thrm:equivalent}:

\begin{theorem}
When given a structure $S$, prediction of a GNN $\mathcal{M}$ defined by Definition~\ref{def:gnn} on $S$ is reducible to a FNN $\mathcal{M}_F$ as defined in definition ~\ref{def:ffnn}.
\label{thrm:equivalent}
\end{theorem}

% \tvn{use the same format as before, say what this theorem does and why it is important. And also say \tool{} relies on this Theorem to do X}
Theorem~\ref{thrm:equivalent} says that every pair of a GNN model $\mathcal{M}$ and any arbitrary input structure $S$ is reducible to an equivalent FNN $\mathcal{M}_F$. 
The full proof for this theorem is available in Appendix~\ref{app:reduction}. Intuitively, \tool{} converts $f_{msg}$, $f_{agg}$, and $f_{upd}$ for each layer sequentially into an equivalent FNN. Since each layer is applied to the input in a compositional manner, this conversion results in an FNN following Lemma~\ref{lem:composition}. 

\subsubsection{Transformation algorithm based on Theorem 1}
Using the recursive transformation in the proof above, we design the GNN-to-FNN transformation algorithm in Fig.~\ref{algo:transform}. Following Theorem~\ref{thrm:equivalent}, the algorithm is proved to be both sound, i.e., the converted FNN $\mathcal{M}_F$ is equivalent to $\langle \mathcal{M}, S \rangle$ and complete, i.e., every GNN model $\mathcal{M}$, given a fixed structure $S$, is convertible into an FNN $\mathcal{M}_F$.

\begin{algorithm}[H]
\small
\SetAlgoLined
\KwIn{Graph Neural Network $\mathcal{M}$ with $k$ layers, Structure $S$}
\KwOut{Equivalent Feed-forward Neural Network $\mathcal{M}_F$}
 $\mathcal{M}_F \leftarrow$ an empty Feed-forward Neural Network\;
 \For{$i \leftarrow 1$ \KwTo $k$}{
  Extract the $i$-th layer of GNN $\mathcal{M}$, denoted as $L_i$\;
  Determine the corresponding FNN layer $F_i$ based on the type of $f_{agg}$ used in $L_i$\;
  \ForEach{node $j \in V_S$}{
      $\mathcal{N}(j) \leftarrow \{j' \in V_S \mid (j', i) \in E_S\}$\;
      \uIf{$f_{agg}$ is mean aggregation}{
       $F_{i, j} \coloneqq f_{upd}\left(\frac{1}{|\mathcal{N}(j)|}\mathbf{1}_{|\mathcal{N}(j)|}^\top \Big\Vert_{k \in \mathcal{N}(j)} f_{msg}(\mathbf{x}_k^{(i)}, \mathbf{x}_j^{(i)}, \mathbf{e}_{kj}) \right)$\;
      }
      \uElseIf{$f_{agg}$ is max aggregation}{
       $F_{i, j} \coloneqq f_{upd}\left(\max_{k \in \mathcal{N}(j)} f_{msg}(\mathbf{x}_k^{(i)}, \mathbf{x}_j^{(i)}, \mathbf{e}_{kj}) \right)$\;
      }
      \uElseIf{$f_{agg}$ is sum aggregation}{
       $F_{i, j} \coloneqq f_{upd}\left(\mathbf{1}_{|\mathcal{N}(j)|}^\top \Big\Vert_{k \in \mathcal{N}(j)} f_{msg}(\mathbf{x}_k^{(i)}, \mathbf{x}_j^{(i)}, \mathbf{e}_{kj}) \right)$\;
      }
  }
  Append $F_i$ to $\mathcal{M}_F$\;
 }
 \caption{Reduction of a Graph Neural Network to a Feed-forward Neural Network. $\mathbf{x}_k^{(i)}$ denote the hidden features of node $k$ at layer $i$ (the output of the previously converted layer), $\Big\Vert$ denotes the stacking operation which stacks each vector $\mathbf{x}_k \in \mathbb{R}^D$ to a form a matrix $\mathbf{X} \in \mathbb{R}^{N\times D}$ where $N$ is the number of stacked vectors. $\mathbf{1}_{|\mathcal{N}(j)|}$ denote the all-1 vector with $\mathcal{N}(j)$ elements.
 %\tvn{why is this algorithm in this Formal setion?}
 }\label{algo:transform}
\end{algorithm}

%\tvn{I still don't understand why this part is here and mixed with the theorems.  May be create a new subsection just for Lemmas and Theorems?}\dat{I have separated Lemmas ,Theorems and algorithm into 2 subsections}
The goal of the transformation algorithm in Fig.~\ref{algo:transform} is to convert each individual message passing layer into the corresponding FNN, using them to construct the equivalent FNN that takes only node and edge features as input with no further concern on the structure. Since $E_S$ is fixed, $\mathcal{N}(j)$ is also fixed for all $j \in V_S$, therefore, the final FNN $\mathcal{M}_F$ only takes as input each node and edge's features.
For more details, see Appendix~\ref{app:transform}.

\paragraph{Example} Using $S_0$ in Fig.~\ref{fig:example_infl_structure}c and the GNN in the BFS example ~\ref{fig:motiv_bfs}. In this case, the next state function is calculated as below for the target node:
\begin{equation}
\mathbf{S}' = \mathbf{X}^{(1)} = \bar{\mathbf{M}} = \left(\begin{array}{l} \max(s_0, s_1) \\ s_1 \end{array}\right)
\end{equation}
The calculation of $\mathbf{S}'$ is an FNN $\mathcal{M}_F$ based purely on the input feature instead of relying on the structure information. Given this equivalent FNN $\mathcal{M}_F$, we can now leverage PROPHECY to infer structure-specific properties.

\subsubsection{Inferring Properties on FNN}\label{subsec:infer_inp_properties}
Given the FNN $\mathcal{M}_F$ transformed from a GNN $\mathcal{M}$ and a structure $S$, we can now infer structure-specific properties using PROPHECY. PROPHECY requires as input \textit{real} execution traces obtained by running the FNN on the training dataset. As a result, we need to collect all instances of the structure $S$ from the original dataset and perform instrumentation on the FNN $\mathcal{M}_F$ to gather its activation patterns and output.

To achieve the above, we make use of Neo4j~\cite{neo4j} database. By encoding the full dataset as Neo4j graphs, we can efficiently retrieve instances of $S$ through graph queries. Since $\mathcal{M}_F$ is constructed from the original GNN through Algorithm~\ref{algo:transform}, we automatically add instrumentation on non-linear layers during this transformation for actual instrumentation.
Particularly, the instrumented points are (1) selected input variables in the $\max$ operation, (2) the activation of $\mbox{ReLU}$, and (3) conditions on the output. Having retrieved all the structure $S$'s instances in the dataset and added instrumentation, we collect the execution traces by running $\mathcal{M}_F$ with all retrieved instances. With these instrumented execution traces, we use PROPHECY to infer the input constraints $\sigma_{inps}$ that imply each output property $P$.

\paragraph{Example.} Consider the converted network based on $S_0$ with the target node $0$. Activation patterns are collected at the calculation of $\max(s_0, s_1)$: we record whether $s_0$ or $s_1$ is chosen in this operation for each instance of $S_0$. If $s_0$ is chosen over node $s_1$, then a condition on the input is $s_0 \geq s_1$ and vice versa. Additionally, output properties on the target node are collected, such as whether $s'_0 = 1$ or $s'_0 = 0$ for each execution. Using the FNN $\mathcal{M}_F$, activation patterns, and output properties, PROPHECY can infer input properties such as $s_0 = 1 \Rightarrow s'_0 = 1$, $s_1 = 1 \Rightarrow s'_0 = 1$, and $s_0 = 0 \land s_1 = 0 \Rightarrow s'_0 = 0$. Given $S_0 = \langle \{0, 1\}, \{(1, 0)\} \rangle$, the structure predicate $\sigma_{S_0\simeq}$ is added to obtain structure-specific properties: $\sigma_{S_0\simeq} \wedge s_0 = 1 \Rightarrow s'_0 = 1$, $\sigma_{S_0\simeq} \wedge s_1 = 1 \Rightarrow s'_0 = 1$, and $\sigma_{S_0\simeq} \wedge s_0 = 0 \land s_1 = 0 \Rightarrow s'_0 = 0$. It is worth noting that in practice, larger networks can result in more complex constraints over the input features.

\subsection{Structure Condition Relaxation}

While structure-specific properties are verifiable, their coverage is limited to a single structure. 
To increase the coverage of a property, we relax the structural constraint from $\sigma_{S\simeq}$ into $\sigma_{S\preceq}$.
% relaxing from graph isomorphism condition into subgraph isomorphism condition.
This means that the input graph structure only has to \emph{contain} (being subgraph isomorphic $\sigma_{S\preceq}$) instead of being \emph{equivalent} (isomorphic $\sigma_{S\simeq}$) to the structure $S$.
Given a structure-specific property $S_P$ in the form of $\sigma_{S\simeq} \land \sigma_{inps} \Rightarrow P$, the corresponding relaxed property $S_{LP}$ is $\sigma_{S\preceq} \land \sigma_{inps} \Rightarrow P$.
Since this relaxation would allow the input graph to have varying numbers of nodes and edges again, the GNN computation can no longer be convertible into an FNN. Thus, the likely property $S_{LP}$ can still be incorrect when the surrounding structure of the input graph changes.

\paragraph{Example} Taking the structure specific property $\sigma_{S_0\simeq} \land s_0 = 0 \land s_1 = 0 \Rightarrow s'_0 = 0$  as input, the likely property is $\sigma_{S_0\preceq} \land s_0 = 0 \land s_1 = 0 \Rightarrow s'_0 = 0$. This resulting likely property can be imprecise.
In detail, for the input graph in the example of Fig.~\ref{fig:example_infl_structure}a which contains $7$ nodes, one $S_0$'s instance which matches this property is the subgraph containing two nodes $2$ and $0$ and one edge from $2$ to $0$. While $s_2 = 0$ and $s_0 = 0$, in this graph, $s'_0 = 1$ instead, this is because of node $0$ has a neighbor $1$ that is already visited.
To improve the precision of likely properties, \tool{} perform dynamic feature property inference in Section.~\ref{sec:dynamic_analysis}.

\subsection{Dynamic Property Inference}\label{sec:dynamic_analysis}
$S_{LP}$ might lose precision due to relaxed structure constraints specifying that the matching graphs only needs to \textit{contain} the original structure $S$. Thus, the matching graphs can also contain additional nodes and edge features that were not originally included in the analysis of $S$ and can influence the prediction of $\mathcal{M}$. 
The goal of this step is to improve the precision of $S_{LP}$ with respect to these unaccounted nodes and edges' features.
% Taking as input a likely property $S_{LP}$, this step outputs an improved dynamic property $S_{DP}$.
We do this by either adding the dynamic input feature predicate $\sigma_{dyna}$ to $S_{LP}$ or replacing the output property $P$ in $S_{LP}$ with a more precise dynamic output feature property $P_{dyna}$.
Both $\sigma_{dyna}$ and $P_{dyna}$ are inferred by leveraging dynamic analysis over a set of aggregated features that take into account the surrounding features of $S$'s instances in the full graph.
% Taking input as a GNN model $\mathcal{M}$, a dataset $\mathcal{D}$ and a likely property $S_{LP}$ of the form $\sigma_{S\preceq} \land \sigma_{inps} \Rightarrow P$,
% the dynamic feature property inference step would extract a set of features and infer a more precise dynamic property $S_{DP}$.
% Using this model, \tool{} construct either the dynamic feature constraint $\sigma_{dyna}$ over the set of features specifying the additional constraints for $P$ to hold, or a more precise dynamic output condition $P_{dyna}$.
% Using $\sigma_{dyna}$ or $P_{dyna}$, \tool{} extends $S_{LP}$ into $S_{DP}$.
% Given $\sigma_{dyna}$, \tool{} extends $S_{LP}$ into   or $P_{dyna}$, extends the likely property $S_{LP}$ into a more precise dynamic feature property $S_{DP}$.
%Choosing which feature to use for dynamic analysis is challenging: The chosen feature set has to give additional information on the input graph in comparison with the existing set of features of node and edge inside $S$ that has already been used to infer structure-specific properties while still being interpretable for analysis and explanation of GNN.

\paragraph{Aggregated Features}
To incorporate the full graph information, we compute the aggregated feature automatically using the GNN's first message-passing layer, which has several advantages in our settings. Firstly, towards the goal of incorporating surrounding information when analyzing GNN, using its own message-passing layer is a straightforward solution. Secondly, since we also want the inferred properties to be lightweight and interpretable, using the first message-passing layer allows the features to have an interpretable format, as we shall show in the example below.
% These features incorporate information from surrounding neighbors for each node in the full graph while still being interpretable, allowing \tool{} effectively to infer interpretable predicates over the input graph.

\tool{} calculates three sets of aggregated features for each node $i$ in the influential structure.
The first set is structural aggregated features $\mathbf{x}^{in}_{i}$, using only the influential structure.
The second set is full-graph aggregated features $\mathbf{x}^{full}_{i}$, using the full input graph.
Finally, the third set is the subtraction $\mathbf{x}^{full}_i - \mathbf{x}^{in}_{i}$, allowing a simple comparison between the structural features and full structure features.
Note that, if multiple message passings on different types of edges are employed, there will be multiple structural features and full-graph features. For these cases, we concatenate all these features for each node.

\paragraph{Inferring dynamic feature properties.} To find dynamic feature properties, \tool{} first trains an interpretable model capturing the deviation between the actual GNN output and the specified output property $P$ over the aggregated features.
Then, \tool{} extracts the dynamic feature properties from these models.
% Depending on whether the graph target a classification or regression problem, \tool{} collects different types of deviations between the actual GNN output and the specified output property $P$ on the dataset $\mathcal{D}$.
The actual type of model and how \tool{} uses the model depends on whether the GNN target a classification or a regression problem.
For classification problems, \tool{} trains a decision tree capturing whether the output property $P$ would hold with respect to the aggregated features.
% Using the actual outcome of the GNN on the full input graphs, \tool{} trains a decision tree over the aggregated features to predict whether the property $P$ would hold. 
Next, \tool{} converts the paths of this decision tree that $P$ holds into the corresponding dynamic feature predicates $\sigma_{dyna}$.
\tool{} uses this predicate $\sigma_{dyna}$ to extend the $S_{LP}$ into dynamic properties $S_{DP}$ taking the form $\sigma_{S\preceq} \land \sigma_{dyna} \land \sigma_{inps} \Rightarrow P$.
For regression problems, \tool{} trains a linear regression model over the aggregated features capturing the difference between the specified output $P$ and the actual output of the GNN.
\tool{} adds the regression model's term with the original output property to retrieve the extended output property $P_{dyna}$ that can also capture these differences.
In this case, the dynamic properties $S_{DP}$ now takes the form $\sigma_{S\preceq} \land \sigma_{inps} \Rightarrow P_{dyna}$.

In summary, the $S_{DP}$ properties, which improve over $S_{LP}$, are defined as:
\begin{equation}
    S_{DP} = \left\{ \begin{array}{ll} \sigma_{S\preceq} \land \sigma_{inps} \land \sigma_{dyna} \Rightarrow P & \text{In classification problem}\\ 
    \sigma_{S\preceq} \land \sigma_{inps} \Rightarrow P_{dyna} & \text{In regression problem} \end{array}\right.
\end{equation}

\paragraph{Example} For the GNN in Fig. ~\ref{fig:motiv_bfs}, given the structure $S_1 = \langle \{1\}, \{\}\rangle$ containing only a single node $1$,
we show how \tool{} calculates the dynamic features and extends the following property $\sigma_{S_1\preceq} \land s_1 = 0 \Rightarrow s'_1 = 0$.
First, \tool{} extracts the structural aggregated feature $x^{in}_1 = \max(s_1) = s_1$ for node $1$ since the structure contains only one node $1$.
\tool{} also extracts the full-graph aggregated feature for node $1$ is $x^{full}_1 = \max_{j \in \mathcal{N}(1)} s_j$.
Thus, the set of dynamic features $f_{dyna_{S_1}}$ on $S_1$ will be: $f_{dyna_{S_1}} = \left(s_1, \max_{j \in \mathcal{N}(1)} s_j, \max_{j \in \mathcal{N}(1)} s_j - s_1\right)$

There are 4 matching instances of the input predicate of the property: the subgraphs containing a single node $0, 2, 3, 4$ respectively in Fig. ~\ref{fig:example_infl_structure}a. 
\tool{} calculates the corresponding $f_{dyna_{S_1}}$ for each of these nodes and obtains the following features $\left(0, 1, 1\right)$, $\left(0, 0, 0\right)$, $\left(0, 1, 1\right)$, $\left(0, 0, 0\right)$.
% Besides the aggregated features, \tool{} collects the information of whether the property holds on each matching instance. 
Following this, \tool{} collects information on whether the property holds for each instance.
The likely property holds on nodes $2, 4$ and does not hold for nodes $0, 3$. %, since, in the next step, node $0$ and $3$ will be predicted as \emph{visited} (i.e., $s'_0 = 1$ and $s'_3$ = 1). 
Using this information, \tool{} then trains a decision tree that predicts on which case the property holds. The learned decision tree on these features would result in the dynamic condition: $f_{dyna_{S_1}, 1} = 1 \Rightarrow holds$ where $f_{dyna_{S_1}, 1} = \max_{j \in \mathcal{N}(1)} = 0$.
Using the decision tree condition, the likely property would be transformed into the following dynamic feature property:
$\sigma_{S_1\preceq}\wedge s_1 = 0 \land \max_{j \in \mathcal{N}(1)}(s_j) = 0 \Rightarrow s'_1 = 0$.
This is the last output of \tool{}.

\section{Evaluation}\label{sec:eval}
%\tvn{why comment out this part?  I think having some description about implementation is appropriate for this section. It also allows the reader to see what libraries or external tools you use, e.g., Z3, PROPHECY, and in/out formats}
%\dat{I think we were out of space}\tvn{don't worry about space now.  We can cut other less interesting parts later}\dat{I uncommented the part}
\tool{} is written in $\sim$20K lines of code in Python. We use Neo4j~\cite{neo4j} for isomorphism checking, and PROPHECY\cite{Gopinath2019} for inferring FNN properties. 
%We also use Z3~\cite{DeMoura2008} as a post-processing step to manually analyze inferred results
% \tvn{do you describe later how you use Z3 to manually check the inferred results?  if not, just remove this sentence}\dat{I removed the sentence}. 
\tool{} takes as inputs a GNN produced using PyTorch~\cite{Pytorch}, and the training dataset $\mathcal{D}$
%\tvn{I thought it also takes in some dataset D as described above?}\dat{I have just updated it},  
and outputs inferred properties of the GNN.

Below we evaluate the effectiveness of \tool{} in discovering GNN properties. We answer 3 research questions (RQs) as follows:
\begin{itemize}
    \item \textbf{RQ1 (Correctness)} evaluates the quality of \tool{}'s resulting properties on both reference and trained GNNs;

    \item \textbf{RQ2 (Applications)} investigates the ability of \tool{}'s inferred properties on detecting backdoor attacks introduced by state-of-the-art methods on GNNs;
    \item \textbf{RQ3 (Efficiency)} measures the efficiency of \tool{}.%\tvn{I would put this as RQ2 and Application as RQ3}\dat{I updated the RQs}

\end{itemize}
% First, in RQ1, we evaluate the quality of \tool{}'s resulting properties on both reference and trained GNNs. In RQ2, we test whether \tool{}'s inferred properties can be used to detect backdoor attacks using state-of-the-art methods on GNNs. In RQ3, we analyze the efficiency of \tool{}. In RQ4 we examine the inferred properties of not well-trained (low accuracy) and also look at the applications of inferred properties.

The experiment results reported here were obtained on an Intel i5-9600K machine%\tvn{what's the CPU speed}
with 6 cores at 3.7GHZ clock speed and 64 GB of RAM running Linux. 

% \begin{mdframed}[roundcorner=10pt,linewidth=1pt,linecolor=black]
% To support reproducibility, we provide the source code and dataset for the experiments\tvn{true?  dataset/results are all provided?}\dat{yes, but the results have to be reproduced using the code and dataset, so I removed it for now (It's too heavy to upload).}\tvn{yes, remove it. You are not required to produce code/data set at submission.  And putting things at an external source is something PL conferences do not like.} at \url{https://doi.org/10.5281/zenodo.8197243}
% \end{mdframed}
%\dat{commented out reproduction package}

\subsection{Benchmarks}\label{subsec:benchmark}
We evaluate GNNInfer on two benchmarks: (1) A benchmark of GNNs simulating classic algorithms and (2) a set of backdoored GNNs on two real-world datasets - Cora and Pubmed from ~\cite{dai2023unnoticeable}.

\paragraph{GNNs simulating classic algorithms.} To provide an initial thorough analysis of \tool{}'s correctness, we construct benchmarks of reference GNNs with identified correct properties and see if \tool{} can re-discover these known properties. Velickovic et al.~\cite{Velickovic2019} have shown that three classic graph algorithms, namely, Breadth-First search (BFS), Depth-First search (DFS), and Bellman-Ford shortest path (B-F), can be converted into message-passing networks with known properties.

All of these 3 tasks are node classification. Taking input as a graph with attributed nodes and edges, all 3 GNNs predict what are the next state (e.g., unvisited, under-visiting, or visited) of each node. The B-F GNN also predicts the next shortest distance from each node toward the original node (we use node $0$ in all our experiments).
We reuse these results to construct the reference GNNs and to check the correctness of \tool{}'s inferred properties.
The number of ground-truth properties for the reference GNN for BFS, DFS, and B-F are 3, 6, and 4 respectively. Details of these properties are explained in Appendix~\ref{app:detailed_problem}.

We also create trained GNNs, which are learned automatically from the data generated by the original algorithm following the prior work~\cite{Velickovic2019}. 
Particularly, we implement 3 reference GNNs along with 33 trained GNNs created from widely used GNN layers i.e., MPNN~\cite{Gilmer2017} and GCN~\cite{Kipf2016}, for benchmark algorithms with various settings as shown in Tab.~\ref{tab:benchmarks}.
The reference GNNs are correct while the other trained GNNs have varying performances as shown in the Appendix~\ref{sec:benchmark_detailed}, including Tabs.~\ref{tab:bfs_dynamic},~\ref{tab:dfs_dynamic} and~\ref{tab:blmfd_dynamic}.

\begin{table}
    \small
    \centering
    \caption{GNN benchmarks. Each target can have different types: classification (\textbf{C}) means GNN predicts discrete categories while regression (\textbf{R}) implies continuous predictions}\label{tab:benchmarks}
    \begin{tabular}{c|c|c|c|c|c}
        \hline
        \multirow{2}{*}{\textbf{Problem}} & \multirow{2}{*}{\textbf{Objective}} & \multirow{2}{*}{\textbf{Type}} & \multirow{2}{*}{\textbf{\#L/\#F}} & \multirow{2}{*}{\textbf{ltype}} & \textbf{Avg. }\\
        & & & & &\textbf{Perf.}\\
        \hline
        BFS & Next state & \textbf{C} & 1-3\textbf{/}2,4,8 & mpnn, gcn & 0.957 \\
        \hline
        DFS & Next state & \textbf{C} & 1-3\textbf{/}2,4,8 & mpnn & 1.0\\
            & Next target& \textbf{C} & 1-3\textbf{/}2,4,8 & mpnn & 0.978\\
        \hline
        B-F & Next state& \textbf{C} & 1-3\textbf{/}2,4,8 & mpnn & 1 \\
             & Next distance& \textbf{R} & 1-3\textbf{/}2,4,8 & mpnn &  $5.47e^{-4}$\\
        \hline
    \end{tabular}
\end{table}

\paragraph{Backdoored GNNs on real-world datasets} To further evaluate the capability of the inferred properties in defending and explaining backdoor behaviors, we employ backdoored GNNs created from the state-of-the-art backdoor attack method UGBA~\cite{dai2023unnoticeable} on two real-world datasets: Cora~\cite{McCallum2000Cora} and Pubmed~\cite{Sen2008Pubmed}. 
Out of three experimented GNNs chosen in UGBA, we experimented on Graph Convolution Networks (GCN)~\cite{Kipf2016}, GraphSAGE~\cite{graphsage} as the Marabou verification tools~\cite{katz2019marabou} does not support quadratic operation in the graph attention layer of GAT. We use the same hyperparameters and training script from UGBA to obtain the trained GNNs.

\subsection{\textbf{RQ1. Correctness of \tool{}}}\label{rq:reference}
We assess \tool{}'s effectiveness using known and trained GNNs (see Tab.~\ref{tab:benchmarks}). The "confidence score", or the likelihood of an inferred property holding for unseen data, is used to measure its performance. In classification tasks, this score is known as proxy accuracy (\textbf{PA}), representing the ratio of the output condition holding over satisfying inputs. For regression tasks, we use proxy error (\textbf{PE}), quantifying the deviation of the output condition from actual model predictions. Properties with \textbf{PA} $> 0.9$ or \textbf{PE} $< 0.01$ are deemed "likely-correct."
We denote \textbf{PA} and \textbf{PE} before and after adding dynamic conditions (explained in Section~\ref{sec:dynamic_analysis}) as $\textbf{PA}_{prior}$, $\textbf{PA}_{full}$ and $\textbf{PE}_{prior}, \textbf{PE}_{full}$ respectively. We quantify the dynamic conditions' impact using the improvement rate (\textbf{IR}) of \textbf{PA} for classification and the reduction rate (\textbf{RR}) of \textbf{PE} for regression tasks. Detailed results for individual GNNs are in Appendix~\ref{sec:benchmark_detailed} with a summary provided below.

\subsubsection{RQ1.1. How correct are the inferred properties on reference GNNs?} 
We collect all the inferred properties for reference GNNs. Among the likely-correct properties, we further conducted a manual assessment to determine whether the inferred properties were equivalent to the desired properties. If the human annotators reached an agreement that a property is indeed equivalent to the desired properties, it was then marked as \textbf{a correct property}. The evaluation process involved two authors independently assessing the properties. If the two annotators disagreed on their decisions, a discussion will be conducted to reach a consensus. 

\paragraph{\textbf{Results}.} \tool{} inferred all 3 known properties for BFS. \tool{} inferred 2 out of 6 known properties for DFS. Additionally, \tool{} inferred 89 likely correct properties for DFS which approximate with the remaining 4 known properties of DFS. \tool{} discovered 3 out of the 4 known properties for B-F. Additionally, \tool{} inferred 5 likely properties for B-F that approximate the remaining one known property of B-F. None of the inferred properties on the reference GNNs has \textbf{PA} lower than 0.9. We give examples for each of these cases below.

\paragraph{\textbf{Case studies}.} For BFS and B-F's next state, the inferred properties correctly indicate 2 expected properties (1) if either there is a visited neighbor or the target node $i$ itself is visited then it will be predicted as visited and (2) If all the surrounding neighbors and the target node $i$ itself are unvisited, then its next state will also be unvisited as shown in the running example (see details in Section~\ref{sec:overview}).

For B-F's next distance, all the generated properties are likely correct with a very small deviation with respect to the ground-truth output conditions (less than $10^{-5}$). For example, one of the inferred properties is: 
    $\sigma_{S_0\preceq} \land d_0 + w_{0, y} - d_y \geq 0 \Rightarrow d'_y = d_y - 0.99 f_{dyna} + 4.29 \times 10^{-6}$
where 
% $f_{dyna} = d_y - \min_{j \in \mathcal{N}(y)}(d_j + w_{j, y}, d_0)$ and $S_0$ is the structure consisting of two nodes $0$, $y$ and an edge between them.
$S_0 = \langle \{1, 0\}, \{(1, 0)\}\rangle$ and $f_{dyna}$ is the offset between the minimum distances inside the structure and in the full input graph towards the target node $0$. This shows that the new distance of node $0$ is the minimum distance from all of its neighbors, with slight deviation due to the linear equation on the output being fit dynamically using features collected from the dataset. 

For DFS, \tool{} captured properties describing when the next status of a node would either remain the same, enter an under-visiting state, and not be the next target. For example, in the structure $S_1 = \langle \{1 \}, \{\} \rangle$ with one node $1$, the properties $\sigma_{S_1\preceq} \land s_{1, c} = 1 \land t_1 = 0 \Rightarrow s'_{1} = c$ for $c \in \{0, 1, 2\}$ correctly described that if node $1$ is not the visiting target, its state remains unchanged.
For the next target objective, given a structure $S_3 = \langle \{0, 1, 2, 3\}, \{(0, 3), (1, 3), (2, 3)\} \rangle$ consisting of 4 nodes $\{0..3\}$ with node $3$ being the target node with all other nodes having forward edges to and backward edges from node $3$, \tool{} inferred the following property $s_{3, 0} = 1 \land s_{1, 1} = 1 \land t_3 = 1 \Rightarrow t'_3 = 0$, which correctly demonstrate that if node $3$ currently is the target and has an under-visiting neighbor, it will not be the next target.
\tool{} infers likely-correct properties when the ground truth condition involves quantifying terms such as there are no ``unvisited'' neighbors or the target node has the highest priority among another node's neighbors. In these cases, \tool{} substitutes these conditions with the conditions of the target's priority being higher or lower than a certain threshold, which is likely correct, but is not precise, on the whole dataset.

\subsubsection{RQ1.2. How correct are the inferred properties on the trained GNNs and the contribution of dynamic analysis}
To assess \tool{}'s performance on trained GNNs where we do not know the ground truth properties, we measure the confidence score (i.e., \textbf{PA} and \textbf{PE}) as well as the corresponding \textbf{IR} and \textbf{RR} on both the BFS, DFS, B-F GNNs and real-world GNNs from~\cite{dai2023unnoticeable} in Tab.~\ref{tab:proxy_metric_summarized} and Tab.~\ref{tab:real_world_gnn_performance}.
\begin{table*}
\small
\caption{Performance of \tool{}'s inferred properties on trained GNNs in BFS, DFS and B-F. Acc. is the accuracy of the models, MSE is the mean squared error of the model. PA is Proxy Accuracy, PE is the proxy error (MSE). IR is the proxy accuracy improvement rate and RR is the proxy error reduction rate}
% \begin{tabular}{lrrrrrrrrrrrrrrr}
% \hline
% \multirow{2}{*}{\textbf{Metric}} & \multicolumn{3}{c}{\textbf{BFS}} & \multicolumn{3}{c}{\textbf{DFS State}} & \multicolumn{3}{c}{\textbf{DFS Target}} & \multicolumn{3}{c}{\textbf{B-F Distance}} & \multicolumn{3}{c}{\textbf{B-F State}} \\\cmidrule(lr){2-4} \cmidrule(lr){5-7} \cmidrule(lr){8-10} \cmidrule(lr){11-13} \cmidrule(lr){14-16}
%  &       \textbf{Acc.} &    \textbf{PA} &    \textbf{IR} & \textbf{Acc.} & \textbf{PA} & \textbf{IR} & \textbf{Acc.} & \textbf{PA} & \textbf{IR} &    \textbf{MSE} & \textbf{PE} & \textbf{RR} &    \textbf{Acc.} & \textbf{PA} & \textbf{IR}  \\
% \midrule
% Mean               &    0.953 & 0.977 & 0.083 &         0.999 &          0.950 &          0.323 &                   0.978 &                    0.934 &                    0.061 &               0.001 &             0.003 &             0.999 &            1.000 &          0.970 &          0.356 \\
% Std. dev &    0.106 & 0.045 & 0.055 &         0.003 &          0.064 &          0.541 &                   0.019 &                    0.047 &                    0.065 &               0.001 &             0.004 &             0.001 &            0.000 &          0.046 &          0.224 \\
% \hline
% \end{tabular}
\begin{tabular}{lrrrrrr}
\hline
\textbf{Method} & \multicolumn{2}{c}{\textbf{Acc.}} & \multicolumn{2}{c}{\textbf{PA}} & \multicolumn{2}{c}{\textbf{IR}} \\ \cmidrule{2-3} \cmidrule{4-5} \cmidrule{6-7} 
 & \textbf{Mean} & \textbf{Std. dev} & \textbf{Mean} & \textbf{Std. dev} & \textbf{Mean} & \textbf{Std. dev} \\ \hline
\textbf{BFS State}&0.953 & 0.106 & 0.977 & 0.045 & 0.083 & 0.055 \\
\hline
\textbf{DFS State} &0.999 & 0.003 & 0.950  & 0.064 & 0.323 & 0.541 \\
\textbf{DFS Target} & 0.978 & 0.019 & 0.934 & 0.047 & 0.061 & 0.065 \\
\hline 
\textbf{B-F State} & 1.000 & 0.000 & 0.970 & 0.046& 0.356 & 0.224 \\
\hline
\end{tabular}
\begin{tabular}{lrrrrrr}
\hline
\textbf{Method} & \multicolumn{2}{c}{\textbf{MSE}} & \multicolumn{2}{c}{\textbf{PE}} & \multicolumn{2}{c}{\textbf{RR}} \\
\hline 
 & \textbf{Mean} & \textbf{Std. dev} & \textbf{Mean} & \textbf{Std. dev} & \textbf{Mean} & \textbf{Std. dev} \\ 
\textbf{B-F Distance} &  0.001 & 0.001 & 0.003 & 0.004 & 0.999 & 0.001\\
\hline
\end{tabular}
\label{tab:proxy_metric_summarized}
\end{table*}

\begin{table} 
\small
\caption{Performance of \tool{}'s inferred properties on real-world GNNs from UGBA~\cite{dai2023unnoticeable}}
\begin{tabular}{l|l|ccc||l|ccc}
\hline
\textbf{Dataset} & \textbf{Model} & \textbf{Acc.} & \textbf{PA} & \textbf{IR} & \textbf{Model} & \textbf{Acc.} & \textbf{PA} & \textbf{IR}\\
\hline
Cora & GCN & 0.7812 & 0.9701 & 1.6453 & GraphSage& 0.7926 &  0.9905 & 0.095 \\
Pubmed & GCN & 0.8539 & 0.7967 & 0.4478 & GraphSage & 0.8523 &  0.9982 & 0.2403 \\
\hline
\end{tabular}
\label{tab:real_world_gnn_performance}
\end{table}

\paragraph{\textbf{Results}.} \tool{} consistently deliverred strong results with a \textbf{PA} of 0.95 (on a scale of 0-1.0) and a low average \textbf{PE} of 0.003 (over the maximum distance of 32 for B-F) across all tasks.
On BFS and DFS Target tasks, \tool{} achieved high \textbf{PA} even prior to employing dynamic feature properties. For DFS state and B-F's objectives, the introduction of dynamic properties led to \textbf{PA} improvements of 0.323, 0.356 and \textbf{PE} reduction of 0.999, demonstrating their effectiveness in capturing varying behaviors of B-F and DFS objectives.

Regarding the real-world GNNs in Tab~\ref{tab:real_world_gnn_performance}, \tool{} achieved high \textbf{PA} (ranging from 0.7967 to 0.9982) and noticeable \textbf{IR} (up to 1.6453 on Cora for GCN), pointing to the potential benefits of dynamic conditions in enhancing generalization in real-world GNNs.

\begin{mdframed}[roundcorner=10pt,linewidth=1pt,linecolor=black]
\textbf{RQ1:} \textit{\tool{}} consistently identified properties with high precision across all GNNs. For reference GNNs, manual assessment shows that, out of 13 ground truth properties, \tool{} inferred 8 correct properties and inferred likely properties that approximate the remaining properties. In trained GNNs, \tool{} demonstrated robust performance with an average \textbf{PA} of 0.95 and an average \textbf{PE} of 0.003. Notably, dynamic analysis contributed to a notable improvement in \textbf{PA} for the DFS state task by 32.3\% and reduced the \textbf{PE} for the B-F distance task by 99.9\%.
%\tvn{put some numbers/stats in this summary} \dat{I have just rewritten it with added stats}
\end{mdframed}

\subsection{RQ2. Defending Against Backdoor Attacks}\label{sec:backdoor}

In this section, we show that \tool{} can be used to infer properties that help effectively defend against backdoor attacks of GNNs. We first provide a brief background on backdoor attacks on GNNs, and then explain how we use \tool{} to effectively defend against them.

Backdoor attacks on GNNs work in three phases: data poisoning, backdoor model training, and backdoored model requirement~\cite{Zhang2021SBA, xi2021graphtrojanattack, dai2023unnoticeable}.
It is commonly assumed that the attacker can access and partially change the dataset of the trained model. This can be done via either publishing the poisoned dataset on an open-source platform, or exploiting the automated data collection process~\cite{ding2023vertexserum}. When other parties use the poisoned dataset to train a prediction model, this results in a backdoored model. When this backdoored model is deployed, the attackers can partially control this model by inserting a trigger. For example, in the sybil (fake user) detection problem~\cite{wang2017sybilscar}, the attacker can manipulate few real users to like specific pages to increase the likelihood of users liking this page being real users. When the Sybil-detecting GNN is trained with this data, it may fail detecting fake users who liked the aforementioned page.  

%\tvn{this opening jumps too quickly and hard to understand.  Need info on what backdoor attacks is and why previous approach is not sufficient.}

We aim at defending against such attack methods. We adopt a widely-used threat model, in which the developers of a GNN have access to both the training set and the clean test set, but do not know about which trigger is inserted, nor which sample is poisoned~\cite{Zhang2021SBA, xi2021graphtrojanattack, dai2023unnoticeable}.
The task is to detect if the GNN is potentially planted with backdoors, and if so, identify the backdoored inputs and defend against them.

Backdoor attacks often follow three characteristics: stealthy, unnoticeable and effective~\cite{dai2023unnoticeable, xi2021graphtrojanattack}. In detail, stealthy means that the backdoor behavior should not decrease the performance of the model on the test set. Unnoticeable means that the number of poisoned samples must be small and finally, effective means that the attack success rate (ASR) has to be high.

We propose to track the properties that warrant these characteristics by using a set of features that we explain below.
Let $P \coloneqq i_{P} \Rightarrow o_{P}$ be a property for a particular structure $S$, where $i_P := \sigma_{S\preceq} \land \sigma_{inps} \land \sigma_{dyna}$ is the input condition and $o_P$ is the output condition. It is important to note here again that $\sigma_{S\preceq}$ is the subgraph isomorphism predicate. Let $\mathcal{D}_{train}$ and $\mathcal{D}_{test}$ be the training and testing set.
Let $SP_{\mathcal{D}_{train}}^{i_{P}}$ and $SP_{\mathcal{D}_{test}}^{i_P}$ be the set of samples that satisfy the input condition $i_P$ on the training set and the test set respectively.
Furthermore, let $SP_{\mathcal{D}_{train}}^{\sigma_{S\preceq}}$ and $SP_{\mathcal{D}_{test}}^{\sigma_{S\preceq}}$ be the set of samples that satisfies the structure condition $\sigma_{S\preceq}$ on the training and test set as well.
Finally, let $SP_{\mathcal{D}_{train}}^{o_P}$, $SP_{\mathcal{D}_{test}}^{o_P}$ be the set of samples that satisfy the output condition $o_P$ on the training and test set.
We show that the backdoor behaviors can be modeled using these features.
To model unnoticeability, meaning that the number of poisoned samples has to be sufficiently small, we hypothesize that the properties that model the behaviors of these poisoned samples would have a small number of supporting samples in the training data as well. We define the support rate as $SPR_{\mathcal{D}_{train}}^{i_{P}} = |SP_{\mathcal{D}_{train}}^{i_{P}}|/ |SP_{\mathcal{D}_{train}}^{\sigma_{S\preceq}}|$.
We choose this rate to be $0.1$ for a property to be unnoticeable.
There may be concerns that being unnoticeable does not mean the property represents a backdoor. 
%In fact, minor classes' example could be the key to improve the performance of the model on the test set. 
Thus, we further strengthen backdoor properties by also modeling stealthiness and effectiveness.
Recall that stealthiness means the backdoored model has to perform well on the test set. 
For the poisoned model to perform well on the test set, there must be much fewer test samples that support the backdoor behavior than on the training set.
We model this behavior by identifying properties that have significantly higher support rate on the training set than on the test set: $SPR_{\mathcal{D}_{test}}^{i_P} /SPR_{\mathcal{D}_{train}} ^ {i_P} < \tau_S$ where $\tau_S$ is a chosen threshold (we set this to 0.05).
We call the properties that match stealthiness and unnoticeability likely-backdoor properties. Likely backdoor properties are considered as backdoor properties once they model the ``effectiveness'' characteristic. 
We model the effectiveness by checking the overriding rate: between a likely backdoor property $P_{likely}$ and a ``benign'' (i.e., unlikely to be backdoor properties) property $P_{benign}$, the overriding rate ($OR$) is defined as the number of samples that model the input conditions in both $P_{likely}$ and $P_{benign}$, but the prediction only satisfy the output condition of $P_{likely}$: $OR_{P_{likely}}^{P_{benign}} = \frac{\left|SP_{\mathcal{D}_{train}}^{i_{P_{likely}} \land i_{P_{benign}} \land o_{P_{likely}}}\right|}{\left|SP_{\mathcal{D}_{train}}^{i_{P_{likely}} \land i_{P_{benign}}}\right|}$

\noindent Recall that the structure condition specifies whether the graph contains a subgraph. Since a graph can contain multiple substructures, a single node can also match multiple input conditions of different properties.
In these cases, if the backdoor properties are effective, their output condition should override the other properties. We choose the likely properties that have a mean overriding rate towards the benign properties $\geq 80\%$ to be the backdoor properties.
After identifying all these backdoor properties, we prune the edges of subgraphs that match these backdoor properties.

We compare the defensive performance of \tool{} versus two state-of-the-art baselines proposed in~\cite{dai2023unnoticeable, chen2022hao, zou2021tdgia}, Prune and Prune+LD. Prune discards edges based on the similarity between the target node and the neighboring nodes' hidden features and Prune+LD additionally discards (isolate) the labels of the nodes of the discarded edges. These homophily-based baselines are widely used in literature~\cite{chen2022hao, zou2021tdgia} for defending against graph backdoors, but have limitations in detecting backdoors planted by UGBA~\cite{dai2023unnoticeable}, which we aim to improve.
\begin{table}
    \small
    \centering
    \caption{Performance of different defensive methods towards UGBA. \textbf{ACC} is the accuracy of the trained GNN on the clean dataset, \textbf{ASR.} is the original attack success rate without any defensive method. For \tool{}, Prune, and Prune-LD, the \textbf{D-ACC} is the accuracy of the trained GNN after the defensive method is added and \textbf{D-ASR} is the attack success rate after the defensive method is applied.
    %\tvn{this table is too wide, making it significantly smaller.  E., use a single decimal point,  Def. Acc = DAC}
    }
    \begin{tabular}{c|ccc|cc|cc|cc}
        \hline
        \multirow{2}{*}{\textbf{Model}} & \multirow{2}{*}{\textbf{Dataset}} & \multirow{2}{*}{\textbf{ACC}} & \multirow{2}{*}{\textbf{ASR}} & \multicolumn{2}{c|}{\textbf{Prune}} & \multicolumn{2}{c|}{\textbf{Prune+LD}} & \multicolumn{2}{c}{\textbf{\tool{}}} \\
        \cmidrule(lr){5-6} \cmidrule(lr){7-8} \cmidrule(lr){9-10}
        & & & & \textbf{D-ACC} & \textbf{D-ASR} & \textbf{D-ACC} & \textbf{D-ASR} & \textbf{D-ACC} & \textbf{D-ASR}  \\
        \hline
        GCN & Cora & 0.781 & 0.996 & 0.782 & 0.991 & 0.781 & 0.993 & 0.774  & \textbf{0.834} \\
        GraphSage & Cora & 0.793 & 0.996 & 0.794 & 0.981 &  0.796 & 0.985 & 0.770 & \textbf{0.959} \\
        GCN       & Pubmed & 0.851 & 0.972 & 0.854 & 0.973 & 0.841 & 0.951 & 0.852 & \textbf{0.914}  \\   
        GraphSage & Pubmed & 0.852 & 0.942 & 0.851 & 0.929 & 0.853 & 0.931 & 0.854 & \textbf{0.893} \\
        \hline
        
    \end{tabular}
    \label{tab:result_defense}
\end{table}

\paragraph{\textbf{Results}.} The result is presented in Tab.~\ref{tab:result_defense}. 
\tool{} consistently outperforms Prune and Prune+LD on both datasets. Particularly, on the Cora dataset and GCN architecture, the success reduction rate of \tool{} is 16.2\%, outperforming Prune (0.5\%) and Prune LD (0.3\%) by more than 30 times. On Cora and with GraphSage, \tool{} reduced the attack success rate by 3.7\% in comparison with 1.5\% and 1.1\% of Prune and Prune+LD respectively.  On the Pubmed dataset, \tool{} reduced the attack success rate by 5.8\% on GCN and 4.9\% on GraphSage in comparison with 2.1\% and 1.1\% of Prune+LD and 0.0\% and 1.3\% of Prune. 
While the coverage of the inferred properties can be limited, leading to limitations in identifying and pruning all backdoor properties, we note that, after performing the pruning, the accuracies of the GNNs on the clean test node on all cases only deviated from the original accuracy by 0.023 at maximum.
This shows that the inferred properties are capable of precisely isolating the backdoor behaviors.
\vspace{5mm}
\begin{mdframed}[roundcorner=10pt,linewidth=1pt,linecolor=black]
\textbf{RQ2:} 
\textit{\tool{}}'s performance in precisely identifying and neutralizing backdoor properties of backdoored GNN without affecting the clean accuracy is promising. In detail, \tool{} outperformed existing baselines by a significant margin: achieving a success reduction rate on the Cora dataset with the GCN architecture that was more than 30-fold higher than the baseline methods (16.2\% vs. 0.5\%/0.3\%) while does not affect the clean test accuracy by more than 0.023 from the original clean test accuracy.
%\tvn{summarize using concrete stats here}\dat{I have just updated it}
\end{mdframed}

%\tvn{tables should always be on top. In other words, you should remove the options htbp and just let LaTeX determines best position}

\subsection{RQ3. Efficiency of \tool{}}\label{sec:efficiency_eval}
% \begin{table}
% \centering
% \caption{Time Analysis of Structure-specific Analysis Stage}
% \label{table:structure-specific-analysis}
% \begin{tabular}{|l|c|}
% \hline
% \textbf{Process} &  \textbf{Max time Taken Per 1,000 samples dataset (seconds)} \\ \hline
% \textbf{Dataset Loading into Neo4j} & 30 \\ \hline
% \textbf{Graph Matching for:} & 1,200 \\ 
% 1-edge structure & 60 \\  
% 2-edges structure & 120 \\  
% 3-edges structure & 420 \\  
% 4-edges structure & 1,200 \\ \hline
% \textbf{GNN to FNN Conversion} & 0.5-1 \\ \hline
% \textbf{Instrumentation} & 120 \\ \hline
% \textbf{Property Inference for} & 300 \\ 
% 1-layered GNN & 60 \\ 
% 2-layered GNN& 70\\ 
% 3-layered GNN& 120 \\
% 4-layered GNN& 160 \\
% 5-layered GNN& 300 \\ \hline
% \textbf{Total} & 1530 \\ \hline
% \end{tabular}
% \end{table}

%\tvn{Could you put some sort of tables or plots in this section? Too much text}
We measure the performance of \tool{} across its three described stages: structure extraction, structure-specific analysis, and dynamic analysis. Since each stage depends on the size of the input graph and the total number of input graphs, we measure the performance of \tool{} with respect to the data volume. Specifically, we choose the benchmark-representative sample size of $1000$, as seen in widely recognized dataset like Cora~\cite{McCallum2000Cora} and Pubmed~\cite{Sen2008Pubmed, dai2023unnoticeable}. We describe the performance of \tool{} in each stage below.

\paragraph{Structural Extraction} This stage consists of performing explanation by GNNExplainer~\cite{Ying2019}, followed by performing mining with gSpan~\cite{yan2002gspan}. 
We measured the total runtime across the experimenting datasets and obtained an average runtime of 2 hours. Most of this runtime is occupied by GNNExplainer (which takes approximately 2 hours). Mining takes only from 10-20 seconds on all datasets. The runtime of GNNExplainer is mostly dedicated to optimizing a mask on each sample and on each prediction GNN, which takes around $7-8$ seconds per graph. 

\paragraph{Structure-specific Analysis} This stage consists of graph matching of influential structures, converting GNN into FNN, running instrumentation, and inferring properties based on the obtained execution traces. This stage takes another 2 hours on our settings. While the GNN-to-FNN conversion is very efficient (taking up only less than 1 second),  performing graph matching and property inference are less efficient. We note that 1,000 input graphs can contain several hundred thousand instances.
For our evaluation, in which the maximum number of retrieved instances was set at 100,000, this retrieval takes up to 20 minutes.
In terms of property inference, since iterative relaxation of properties requires verification, this takes up to 90 minutes.

\paragraph{Dynamic Analysis} Dynamic analysis only concerns the first layer of GNN, so run time is nearly constant given the number of samples. In our case, this takes less than 90 seconds for all properties.

\paragraph{Backdoor defense} After inferring properties, we can optionally include a backdoor defense stage (see Section~\ref{sec:backdoor}). The backdoor defense consists of identifying the backdoor properties and performing pruning of these properties. The backdoor properties identification takes $2$ minutes in our settings for each test set and for each structure, given that we obtained the training support set from the structure-specific analysis stage. 
Since pruning requires performing graph-isomorphism checking again and pruning the structure having matched features, it takes up to 30 minutes for the largest structure containing 3 nodes and 7 edges. In our experiments, this takes a total of 2 hours.

\vspace{2cm}
\begin{mdframed}[roundcorner=10pt,linewidth=1pt,linecolor=black]
\textbf{RQ3:} The inference process of \tool{} takes approximately $4$ hours in total as upper bound in our experiment. While GNN-to-FNN conversion is very efficient (takes less than one second), other steps such as explanation, graph matching, and property inference are less efficient. While there is room to improve efficiency, we believe that the current efficiency of \tool{} is reasonable for offline processing.
%The explanation step takes 2 hours due to the reliance on GNNExplainer, an optimization-based approach that is required to precisely obtain the influential structure for every single node in all graphs in the dataset. The optional backdoor defense mechanism takes an additional 2 hours. 
%\tvn{add one sentence here saying something about this timing is reasonable because ..  Otherwise it might appear to be very slow, e.g., 4 hrs}\dat{I have just updated it}
\end{mdframed}

\section{Discussion}\label{sec:discussion}
% Dat: TRYING TO Put a more interesting discussion here, if not, will skip.
In this section, we discuss how \tool{} can be used for debugging GNN models given certain expected properties, enabling a property-based approach to GNN debugging.
%\tvn{I am not sure if we want this section.  Either extend it and put it as a case study or remove it entirely.}
% We devote this section to discussing additional potential applications of \tool{}'s, particularly, on debugging DNN models. If we have expected model properties, we show that it would be possible to perform a property-based debugging of GNNs with \tool{}'s inferred properties.

% inferred properties based on current literature as well as an in-depth analysis of the inferred properties. 

%A common practice in debugging deep learning frameworks is differential testing~\cite{Deng2022DeepRel, Gu2022muffin, wei2022freefuzz}. In these works, the same inputs are provided to multiple frameworks that are designed to perform the same task. These techniques can reveal faults by analyzing the mismatch between the outputs of these frameworks. However, it is still hard to apply differential testing to debug DNNs due to the heterogeneity in the construction of DNNs~\cite{Humbatova2020taxnonomyofrealfault, Amershi2019, islam2019comprehensivednnbugs}: if a model fails at the final stage, it is unclear whether it is because of the model structure, weights, or dataset. While there have been prior works leveraging properties and verification to debug neural networks~\cite{Sun2022Care, usman2021nnrepair, LPAR23:Minimal_Modificatios'_of_Deep}, these works are not applicable to GNN due to the same varying structure challenge.

To demonstrate this idea, we conduct a small analysis of the three problems - BFS, DFS, and B-F and show how the inferred properties can be used in conjunction with the expected properties to debug GNN models.
Firstly, if \tool{} infers high-confident properties that follow the desired property, the developers can be more confident that the analyzed GNN is correct. As an example, the following inferred properties with high confidence score on $\text{dfs}_1$ in Section~\ref{rq:reference}:
$\sigma_{S_1\preceq} \land s_{1, 0} = 1 \wedge t_1 = 0 \Rightarrow s'_1 = 0$, shows that this GNN is likely to model well the property that if a node is unvisited and it is not the target, it will stay unvisited.  
Secondly, if \tool{} discovers a property having high confidence score but contradicts an expected property, then these properties can help highlight the GNNs' error. For instance, consider the GNN  $\text{bfs}_1$ in Tab. ~\ref{tab:bfs_dynamic}. We see a property $\sigma_{S_0\preceq} \land s_1 = 0 \land s_0 = 1 \Rightarrow s'_0 = 0$ with a perfect confidence score, but is wrong with respect to the expected behavior: $\text{bfs}_1$ incorrectly predicts a node as unvisited even if it is already visited, given the presence of an unvisited neighbor. 
Finally, we can also identify GNNs' error by finding the properties that agree with the ground truth but having low confidence scores, such as two inferred properties of $\text{bfs}_5$ from Tab. ~\ref{tab:bfs_dynamic}: (i) $\sigma_{S_1\simeq}\land s_1 = 0 \land \max_{j \in \mathcal{N}(1)} s_j = 0 \Rightarrow s'_1 = 0$, and (ii) $\sigma_{S_0\preceq} \land s_1 = 0 \wedge s_0 = 1 \Rightarrow s'_0 = 1$. These properties together suggest that the GNN probably fails in modeling the case where the node and its surrounding neighbors are all unvisited, or the transition from unvisited to visited of one node.

These analyses hint that it is possible to use inferred properties to ensure GNN's correctness as well as highlight its errors if we are given expected properties. Since most of the current machine learning problems do not come with expected properties, an interesting future direction would be applying \tool{} to different models to infer a set of common expected properties that every future-developed model should have. 

\section{Related Work}\label{sec:related}
% \tool{} leverages ideas and tools from property inference (using PROPHECY), verification (e.g., using a verifier to check inferred properties),  and the explanation of GNNs (using GNNExplainer). 

% \textbf{Neural Network Verification}
% For example, Marabou, Reluplex\cite{Katz2019, katz2017reluplex} use simplex-based solvers while DLV~\cite{huang2017safety} and Planet~\cite{ehlers2017formal} use existing SMT solvers to verify neural networks' properties; Branch and Bound~\cite{Bunel2019}, Split and Conquer~\cite{wu2020parallelization} are popular parallel algorithms used by these approaches to scale verification. 
% ERAN~\cite{gpupoly, Singh2019, Singh2019boosting}, ReluVal~\cite{wang2018formal} and AI2~\cite{Gehr} are abstraction-based verifiers. They use abstract interpretations along with different domains (e.g., interval, zonotope, polytope, etc.) or MILP solvers to improve the verification precision.

% 
\paragraph{Neural Network Analysis}
The line of work that is closely related to \tool{} is inferring properties~\cite{Gopinath2019, nguyen2022gnninfer, geng2023towards}, verification of properties~\cite{katz2017reluplex, Katz2019, huang2017safety, ehlers2017formal} and repairing networks with properties~\cite{Sun2022Care, usman2021nnrepair,LPAR23:Minimal_Modifications_of_Deep} for DNN.
PROPHECY~\cite{Gopinath2019} proposed the usage of using activation patterns of ReLU \cite{relu} and max-pool \cite{AlexNet} to infer DNN's properties in the form of an input condition implying an output condition, both in the form of linear inequalities.
\cite{geng2023towards} proposed a mining-based approach using activation patterns to infer DNN properties and apply them as neural network specifications.
Both \cite{Gopinath2019, geng2023towards} can only infer FNN and CNN properties and cannot be applied to infer GNN properties, which \tool{} aims to address. 
\tool{}'s method to debug GNN also differs from existing property-based DNN debugging works~\cite{Sun2022Care, usman2021nnrepair, LPAR23:Minimal_Modifications_of_Deep}. 
These works try searching for the weights that let the DNN model a set of expected properties. 
They ensure the correctness of the new weights using the DNN verification techniques~\cite{katz2017reluplex, Katz2019, huang2017safety, ehlers2017formal, katz2019marabou, wang2018formal, Gehr, gpupoly, Singh2019}. Since the used verification techniques have yet to support GNNs due to the same varying structure problem, these debugging techniques~\cite{Sun2022Care, usman2021nnrepair, LPAR23:Minimal_Modifications_of_Deep} cannot be applied on GNNs either.
\tool{} instead uses the inferred properties as filters to remove backdoored input. Finally, we note that \tool{} is the first fully developed, implemented, and thoroughly evaluated tool that analyzes GNNs through FNN conversion. Recent work~\cite{nguyen2022gnninfer} gives an outline of ideas about GNN analysis via FNN. However, it only provides sketch ideas without any algorithms, formal analyses, implementations, applications on backdoor defense, or evaluations.

\paragraph{Backdoor attacks on Graph Neural Networks} 
Backdoor attacks against deep learning model in general is a specific type of data poisoning. Backdoor attacks work in two phases: the data collection phase and the training phase. In the data collection phase, the attacker poisons the training dataset by attaching triggers to a set of samples with target labels. Training the model on this dataset would result in a backdoored model that associates the presence of the trigger to the target label. This gives the attacker partial control over the model's output. 
SBA~\cite{Zhang2021SBA} initiates backdoor attacks on GNN by attaching a fixed graph to every poisoned sample. GTA~\cite{xi2021graphtrojanattack} adopts an adaptive trigger generation via bi-level optimization and UGBA further increases the efficiency of backdoor attach by efficiently choosing representative nodes to attach the trigger. While SBA and GTA's attack can be defended by using the homophily property graph neural networks, namely, with Prune and Prune+LD in \cite{dai2023unnoticeable}, none of these methods are effective against UGBA. 
Our framework can also be seen as a tool to analyze and defend against backdoor attacks, by detecting the properties that match the backdoor characteristics, we can defend against these properties by removing subgraphs that model these properties.
CARE~\cite{Sun2022Care} also tries applying particle swarm optimization to adjust neural network weights to defend against backdoor attacks. However, CARE requires to know the poisoned samples beforehand, which might not be practical. On the other hand, \tool{} infers the backdoor properties automatically without these restrictions.

\paragraph{Explaining Graph Neural Networks}
Our work can also be regarded as a different way to aid understanding GNNs. In this line of work, ~\cite{ying2019gnnexplainer} introduced a method using meta-gradient to optimize a mask of the important score for each node, edge, and feature, \cite{Sahoo2020} proposes encoding the problem of identifying important features using an SMT formulation and solves it using SMT solver, \cite{Huang2020} proposes a framework to compute the most significant features and ~\cite{Vu2020a} explains GNN by inferring a probabilistic graphical model and PGExplainer ~\cite{luo2020parameterized} learns a model predicting the important score of each node and edge.
While these works aim to directly attribute which part of the input is responsible for the output to help human experts gain intuition to debug the GNNs, \tool{} outputs properties. These properties can represent a class of correctness or faulty inputs as seen in Section~\ref{sec:discussion}.
Since PGM Explainer~\cite{Vu2020a} also outputs a probabilistic graphical model that can be regarded as a probabilistic formula relating the input and output of the model, it would be interesting to see additional work built upon PGM Explainer for automated debugging of GNNs in future works.

%\tool{}'s likely properties have high confidence and can give a general explanation of the behavior similar to that of ~\cite{Vu2020a}, the difference is while 

\section{Conclusion And Future Work}\label{sec:conclusion}
We introduced \tool{}, a property inference tool for GNN. The foundation of \tool{} is the formal proof and the algorithm to reduce a graph neural network with a specific structure into an equivalent FNN. Based on this, \tool{} infers the GNN's property on the given specific structure by using property inference tool on the reduced FNN. Finally, \tool{} generalizes the inferred structure-specific properties to a class of input graph that contain the chosen structure with dynamic analysis. Experiment results demonstrated that \tool{} can infer correct and highly confident properties on both synthetic as well as real-world GNNs. Furthermore, we also used \tool{}'s inferred properties to detect and defend against the state-of-the-art backdoor attack, namely UGBA~\cite{dai2023unnoticeable}. Experiments showed that \tool{} increases the defense rate up to 30 times in comparison with the two defense baselines. For future works, we plan to improve \tool{} and use it for other tasks, e.g., verification of GNNs.
%\tvn{either add a bit more details, or remove this. I would just remove this para}

\bibliography{sample-base}
\newpage

%% Appendix
\appendix

\section{Appendix} \label{sec:appendix}
\subsection{Detailed Problem Settings}\label{app:detailed_problem}
\subsubsection{BFS} \label{subsec:appen_bfs_problem}
The BFS GNN takes as input a graph in which each node $i$ has a state feature $s_i$, $s_i = 0$ if this node is unvisited and $s_i = 1$ if this node is visited. Given that a set of node is initially visited, this GNN would predict the state of each node in the next visiting iteration.

\noindent\textbf{Groundtruth property}
As stated in~\cite{Velickovic2019}, it is desired that GNNs follow the following properties: 
\begin{enumerate}
    \item If a node $i$ is visited, then it will continue to be \textit{visited}: $s_i = 1 \Rightarrow s'_i = 1$
    \item If a node $i$ has a visited neighbor $j$, in the next step, it will be visited $\exists j \text{ s.t. } (j, i) \in E \land s_j = 1 \Rightarrow s'_i = 1$.
    \item If both $i$ and its surrounding neighbors are unvisited, then it will continue to be visited $\forall j \in \mathcal{N}(i), s_j = 0 \wedge s_i = 0 \Rightarrow s'_i = 0$.
\end{enumerate}

\textbf{Reference GNN} We design a GNN model which is guaranteed to behave according to the desired properties. Particularly, we construct the reference GNN for the Parallel BFS problem based on~\cite{Velickovic2019}, taking as input a graph $G = (V_G, E_G)$
\begin{align}
		m_{j, i} &= f_{msg}(s_j, s_i) = s_j \\
		\bar{m}_i &= f_{agg}(m_{j, i} | (j, i) \in E_G) = \max\limits_{j \in \mathcal{N}(i)} m_{j, i}\\
		s'_i &= f_{upd}(\bar{m_i}, s_i) =  \bar{m}_i
\end{align}
This would result in $y_i = \max\limits_{j \in \mathcal{N}(i) \cup \{i\}} s_j$. This relation between the input $s_j$

\subsubsection{DFS}\label{subsec:appen_dfs_problem}
The DFS GNN takes as input also a graph, with each node $i$ having two features $\left(\begin{array}{l} s_i\\ t_i \end{array} \right)$. $s_i = 0$ if the node is unvisited, 1 if undervisiting and 2 if visited. The additional feature $t_i$ is used to determine which exact node is underwriting. At each step, the DFS would predict (1) which node to be undervisiting next $t'_i$ and the next state $s'_i$ of all nodes.
The next underviting node will have feature $t'_i = 1$, while the rest have $t'_i = 0$.
To describe DFS as a message passing procedure, we present the algorithm for 2 objectives in Fig.~\ref{fig:dfs_algo_state}, Fig.~\ref{fig:dfs_algo_target}.

% For DFS, we have three known properties: (1)
% If the node is not the target, then the output class should be the same: $t_i = 0 \wedge s_{i, 0} = 1 \Rightarrow s'_i = 0$, $t_i = 0 \wedge s_{i, 1} = 1 \Rightarrow s'_i = 1$, $t_i = 0 \wedge s_{i, 2} = 1 \Rightarrow s'_i = 2$. (2) If the node is the target, then if there exists a forward \emph{unvisited} neighbor, the next state will be \emph{under-visiting} $t_i = 1 \wedge (\max_{j \in \mathcal{N}_f(i)} s_{j, 0}) = 1 \Rightarrow s'_i = 1$ and (3) Otherwise, next state will be \emph{visited} $t_i = 1 \wedge (\max_{j \in \mathcal{N}_f(i)} s_{j, 0} = 0) \Rightarrow s'_i = 2$

\textbf{Algorithm} We describe in detail the algorithm for each objective \emph{next state} and \emph{next visiting target} in Algorithm ~\ref{fig:dfs_algo_state} and Algorithm ~\ref{fig:dfs_algo_target}, respectively.

\begin{algorithm}[H]
	\SetAlgoLined
	\SetKwInOut{Input}{input}\SetKwInOut{Output}{output}
	\Input{Target node $t$, States map $S$, node $i$}
	\Output{The new state $s'_i$ for node $i$}

	\lIf{$i \neq t$}{\Return $S[i]$}
	Unvisited neighbors set $U_{i, f} \gets \{j \in \mathcal{N}_f(i)| S[j] = 0\}$\;
	\lIf{$U_{i, f} \neq \emptyset$}{\Return 1}
	\lElse {\Return 2}
        
	\caption{Determine the next state for node $i$}
	\label{fig:dfs_algo_state}
\end{algorithm}

\begin{algorithm}[H]
	\SetAlgoLined
	\SetKwInOut{Input}{input}\SetKwInOut{Output}{output}
	\Input{Target node $t$, States map $S$, priority function $P$, Target node next state $s'_t$}
	\Output{Next visiting target node $nt$}

	\If{$s'_t = 1$}{
		\Return $\arg\min_{j \in \mathcal{N}_f(t)| S[j] = 0} P(j)$\;
	}
	\Else{
		under-visiting node set $V_t \gets \{j \in \mathcal{N}_b(t) | S[j] = 1 \land j \neq t\}$\;
		\If{$V_t = \emptyset$}{
			\Return -1\;
		}
		\Return $\arg\max_{j \in V_t} P(j)$
	}
	\caption{Determine the next visiting target}
	\label{fig:dfs_algo_target}
\end{algorithm}

Where $N_f(i)$ and $N_b(i)$ identify all the forwards and backwards neighbors of node $i$ respectively, finally, given the new target node $nt$, the edge from the current target node $t$ to $nt$ will be added to the \emph{backward} edge: $E_G = E_G \cup \{(nt, t)\}$, $\lambda_G = \lambda_G \cup \{(nt, t)\mapsto 1\}$. 
Note that the GNN only makes two predictions on node-level and the input graph's edges are already added in the dataset at each step.

\textbf{Ground truth property} With respect to the next state output:
\begin{itemize}
    \item If the node is not the target, then the output class should be the same: $t_i = 0 \wedge s_{i, 0} = 1 \Rightarrow s'_i = 0$, $t_i = 0 \wedge s_{i, 1} = 1 \Rightarrow s'_i = 1$, $t_i = 0 \wedge s_{i, 2} = 1 \Rightarrow s'_i = 2$.
    \item If the node is the target, then if there exists a forward \emph{unvisited} neighbor, the next state will be \emph{under-visiting} $t_i = 1 \wedge (\max_{j \in \mathcal{N}_f(i)} s_{j, 0}) = 1 \Rightarrow s'_i = 1$
    \item Otherwise, next state will be \emph{visited} $t_i = 1 \wedge (\max_{j \in \mathcal{N}_f(i)} s_{j, 0} = 0) \Rightarrow s'_i = 2$
\end{itemize}
With respect to the next target determination:
\begin{itemize}
    \item If the node is the target, then it will not be the next target $t_i = 0 \Rightarrow t'_i = 0$
    \item If the node is the target's smallest priority \emph{unvisited} forward neighbor, then it will be the next target. $s_{i, 0} = 1 \wedge t_i = 0 \exists j \in \mathcal{N}_b(i) \wedge t_j = 1 \wedge p_i = \min_{i' \in \mathcal{N}_f(j)} p_{i'} \Rightarrow t_i = 1$
    \item If the node is the target's largest priority \emph{under-visiting} backward neighbor, and the target node does not have any forward \emph{visited} neighborhood, then it will be the next target. $s_{i, 1} = 1 \wedge t_i = 0 \exists j \in \mathcal{N}_f(i) \wedge t_j = 1 \wedge p_i = \max_{i' \in \mathcal{N}_b(j)} p_{i'} \wedge \max_{i' \in \mathcal{N}_f(j)} s_{i', 0} = 0 \Rightarrow t_i = 1$
\end{itemize}

\textbf{Correct Reference GNN} 
\begin{equation}
s'_i = \left\{ \begin{array}{ll} s_i & \text{ if } t_i  = 0 \\ 1 & \text{ if } t_i = 1 \wedge \max\limits_{j \in \mathcal{N}(i) \setminus \{i\}} s_{j, 0} = 1\\ 2 & \text{ otherwise }\end{array} \right.
\end{equation}
To derive the message passing form of $s'_i$, we split the calculation of its values into 3 cases: $s'_{i, 0}, s'_{i, 1}, s'_{i, 2}$ corresponding to node $i$ next state to be predicted as \textit{unvisited}, \textit{undervisiting} and \textit{visiteds}. $s'_i$'s value is determined by $s'_i = \arg\max_j s'_{i, j}$
and given $C$ is a large number (we choose $100$ in our implementation).
\begin{itemize}
    \item $s'_{i, 0} = \mbox{ReLU}(- C  t_i + s_{i, 0})$ If node $i$ is not the target node, the value of $s'_{i, 0}$ will be $0$, otherwise, it will be the former value $s_{i, 0}$.
    \item $s'_{i, 1} = \mbox{ReLU}\left(- C  \left((1 - t_i) + \min_{j \in \mathcal{N}_f(i)}\left(1 - s_{j, 0} + t_j\right)\right) + 1\right) + \mbox{ReLU}(- C  t_i + s_{i, 1})$. If node $i$ is the target node (i.e., $t_i = 1$, then the first term will be reduced to the indication of whether there exists a node $j$ different from node $i$ such that node $j$ current state is \emph{visited}. Otherwise, if $t_i = 0$, $s'_{i,1}$ will be reduced to $s_{i, 1}$
		\item $s'_{i, 2} = 1 - s'_{i, 0} - s'_{i, 1}$
\end{itemize}
The $\min$ terms in $s'_{i, 1}$ can be straight-forwardly implemented using the message passing mechanism, for example, for $s'_{i, 1}$:
\begin{align}
    m_{j \rightarrow i} &= \mbox{ReLU}(1 - s_{j, 0} + t_j)\\
    \bar{m}_i &= \min_{j \in \mathcal{N}_f(i)} m_{j, i}\\
    s'_{i, 1} &= \mbox{ReLU}\left( -C  \left((1 - t_i) + \bar{m}_i\right)\right) + 1)+ \mbox{ReLU}(-C  t_i + s_{i, 1})
\end{align}

The next target node objective is rather challenging since GNNs infer node-level classification has to infer the result based solely on the node's updated hidden feature (i.e., the next target node has to know that it is the highest-priority forward \emph{visited} neighbor of a target node or the highest-priority backward \emph{under-visiting} neighbor.)

Since the GNN compute node's next hidden features based solely on the current neighbor's feature, we use a message passing mechanism to save the aforementioned top-priority forward \text{unvisited} ($tfu$) neighbor and top-priority backward \text{undervisiting} ($tbu$) neighbor and another message passing procedure so that corresponding neighbor acknowledges that it is the next target node.

For any node $i$, the $lfu$ can be calculated using the following equation:
\begin{align}
	lfu_i &= \mbox{ReLU}(\max_{j \in \mathcal{N}_f(i)} (p_j - C(1 - s_{j, 0} + t_j)\\
	lbu_i &= \mbox{ReLU}(\max_{j \in \mathcal{N}_b(i)} (p_j - C(1 - s_{j, 1} + t_j)) - C \cdot lfu_i
\end{align}
$lfu_i$ and $lbu_i$ is equal to the maximum priority of the node $i$'s forward \emph{visited} and backward \emph{undervisiting} neighbors respectively. The term $t_j$ ensures that $j$ is not the current target node.
\begin{equation}
    t'_i = \left(\begin{array}{ll}
    &\mbox{ReLU}\max_{j \in \mathcal{N}_b(i)} \left(
    \begin{array}{ll}
    1 -\\
    C (1 - t_j + \mbox{ReLU}(lfu_j - p_i) + \\
    \mbox{ReLU}(p_i - lfu_j)) - C \cdot t_i
    \end{array}\right)+  \\
    &\mbox{ReLU}\max_{j \in \mathcal{N}_f(i)}\left(
    \begin{array}{ll}1 - \\C (1 - t_j + \mbox{ReLU}(lbu_j - p_i) +\\ \mbox{ReLU}(p_i - lbu_j)
    \end{array}\right) -\\ &C\cdot t_i)\\
    \end{array}\right)
\end{equation}

\subsubsection{Bellman-Ford}\label{subsec:appen_blmfd_problem}
Bellman-Ford (B-F) GNN takes as input a graph where each node is assigned with feature $x_i = \left(\begin{array}{c} s_i \\ d_i \end{array}\right)$ where $s_i \in \{0, 1\}$ is the current state of the node (visited or unvisited) and $d_i$ is the current shortest distance from the node to the root node (node $0$) and each edge is associated with a weight feature $w_{ji}$, denoting the weight of the edge $(j, i) \in E$.

\textbf{Ground truth property} There exists 2 ground-truth properties regarding the node $i$'s next updated distance and also its visited state. Regarding node $i$'s visited state:
\begin{enumerate}
    \item If all the surrounding node of $i$ is \emph{visited} and node $i$ is also \emph{visited} then node $i$ will be marked as \emph{visited}: $s_i = 0 \wedge \max_{j \in \mathcal{N}(i)} = 0 \Rightarrow s'_i = 0$. The structure predicate, in this case, consists of a node $i$, and its out-of-structure neighbors' features $\max_{j \in \mathcal{N}(i)}$.
		\item If node $i$ is \emph{visited}, then it will continue to be predicted as \emph{visited}: $s_i = 1 \Rightarrow s'_i = 1$. In this case, the structure predicate only consists of a single node $i$.
		\item If there exists a node $j$ in the node $i$'s neighbor such that node $j$ state is \emph{visited}, then node $i$ will be marked as \emph{visited} $\max_{j \in \mathcal{N}(i)} = 1 \Rightarrow s'_i = 1$. In this case, the structure predicate consists of a node $i$ and a node $j$ which the state is \emph{visited}.
\end{enumerate}
Regarding node $i$ new distance: node $i$'s updated distance is the minimum between its current distance and the sum of each neighbor $j$'s distance and weight toward $i$: $d'_i = \min_(\min_{j \in \mathcal{N}(i)} d_j + w_{j, i}, d_i)$. The structure predicate may consist of 1 node when its new distance remains the same and 2 nodes or when the new distance is updated.

\textbf{Reference GNN} The reference GNN is straightforward, comprising of 2 message passing operations: the reused message passing operation from Parallel BFS and the following message passing operation for calculating $d'_i$. Given that the GNN add self-loops (which is a common setting in the implementation of GNN) for each node:
\begin{align}
    m_{j, i} &= d_{j} + w_{j, i}\\
    \bar{m}_{i} &= \min_{j \in \mathcal{N}(i)} m_{j, i}\\
    d'_i &= \bar{m}_i\\
\end{align}

\section{Reduction}\label{app:reduction}

We give detailed proof of each lemma and theorem below.

\begin{proof} (Lemma~\ref{lem:combine})
Let $L^{(k)}_{i}$ be  the $k$th layer of $\mathcal{M}_{i}$, $V^{(l)}_{i}$ and $V^{(l)}_{i+1}$ be $L^{(k)}_{i}$'s input and output variable sets, respectively. Let $U^{(k)} = \bigcup_{i=1}^{n} V^{(k)}_{i}$ be the union of output variable sets $V^{(k)}_{i}$ with $1 \leq i \leq n$ and $n_{k}$ be the number of elements of $U^{(k)}$. 

Consider the relation $L^{(k)}$ between $U^{(k-1)}$ and $U^{(k)}$ are combination of $k^{th}$ layers of every neural network $\mathcal{M}_{1}, \ldots, \mathcal{M}_{n}$ with $1 \leq k \geq l$:

\begin{equation}
    L^{(k)}(x^{(k-1)}_{1}, \ldots ,x^{(k-1)}_{n_{k-1}}) = (x^{(k)}_{1}, \ldots ,x^{(k)}_{n_{k}})
\end{equation}
where, $x^{(j)}_{1}, \ldots ,x^{(j)}_{n_{j}}$ are all of distinct variables in $U^{(j)}$. 

We can prove that each relation is actually a function by proving each valuation of $(x^{(k-1)}_{1}, \ldots ,x^{(k-1)}_{n_{k-1}})$ are assigned to one and only one valuation of $(x^{(k)}_{1}, \ldots ,x^{(k)}_{n_{k}})$. 

Indeed, let us assume that there exists a two valuation of $(x^{(k)}_{1}, \ldots ,x^{(k)}_{n_{k}})$:  $(\varv(x^{(k)}_{1}), \ldots ,\varv(x^{(k)}_{n_{k}}))$ and $(\varv'(x^{(k)}_{1}), \ldots ,\varv'(x^{(k)}_{n_{k}}))$ such that:
\begin{equation}
    L^{(k)}(\varv(x^{(k-1)}_{1}), \ldots ,\varv(x^{(k-1)}_{1})) = (\varv(x^{(k)}_{1}), \ldots ,\varv(x^{(k)}_{n_{k}}))
\end{equation}
\begin{equation}
    L^{(k)}(\varv(x^{(k-1)}_{1}), \ldots ,\varv(x^{(k-1)}_{1})) = (\varv'(x^{(k)}_{1}), \ldots ,\varv'(x^{(k)}_{n_{k}}))
\end{equation}

\begin{equation}
    (\varv(x^{(k)}_{1}), \ldots ,\varv(x^{(k)}_{n_{k}})) \neq (\varv'(x^{(k)}_{1}), \ldots ,\varv'(x^{(k)}_{n_{k}}))
    \label{eq:lmeq1}
\end{equation}

Equation~\ref{eq:lmeq1} implies that there exists at least one variable $x^{(k)}_{*}$ of $U^{(k)}$:
\begin{equation}
    \varv(x^{(k)}_{*}) \neq \varv'(x^{(k)}_{*})
    \label{eq:lmeq2}
\end{equation}

Without loss of generality, we can assume that $x^{(k)}_{*}$ is $x^{(k)}_{1}$. Since $\mathcal{M}_{1}, \ldots, \mathcal{M}_{n}$ are different, meaning that their variable sets $V^{(k)}_{1}, \ldots, V^{(k)}_{n}$ are distinct. This implies that $x^{(k)}_{1}$ belongs to one and only one variable set $V^{(k)}_{*}$ of layer $L^{(k)}_{*}$ of neural network $\mathcal{M}_{*}$.

We also have the input variable set of $L^{(k)}_{*}$ is a subset of $U^{(k-1)}$. Without loss of generality, we suppose that $x^{(k-1)}_{1}, \ldots, x^{(k-1)}_{|V^{(k-1)}_{*}|}$ ($|V^{(k-1)}_{*}|$ is the number of elements of $V^{(k-1)}_{*}$) are all of the input variables of $L^{(k)}_{*}$. We have:

\begin{equation}
    \varv(x^{(k)}_{1}) = L^{(k)}_{*}(
    \varv(x^{(k-1)}_{1}), \ldots, \varv(x^{(k-1)}_{|V^{(k-1)}_{*}|}))
    \label{eq:lmeq3}
\end{equation}

\begin{equation}
    \varv'(x^{(k)}_{1}) = L^{(k)}_{*}(
    \varv(x^{(k-1)}_{1}), \ldots, \varv(x^{(k-1)}_{|V^{(k-1)}_{*}|}))
    \label{eq:lmeq4}
\end{equation}

From Equation~\ref{eq:lmeq2},~\ref{eq:lmeq3} and~\ref{eq:lmeq4}, we can see that $L^{(k)}_{*}$ can output different values with the same input and thus a contradiction because $L^{(k)}_{*}$ is a function. 

Thus, we now can conclude that $L^{(k)}$ is a function. 

Moreover, as $L^{(k)}$ is a combination of $L^{(k)}_{1}, \ldots L^{(k)}_{m}$, $L^{(k)}$ is one of the following: (1) affine transformation, (2) the RELU activation function (3) a max pooling function. 

Therefore, following Definition~\ref{def:ffnn}, we can conclude that the combination of $\mathcal{M}_{1}, \ldots, \mathcal{M}_{n}$, which is a composition of $L$ functions  $L^{(1)}, \ldots L^{(l)}$ is a feed-forward neural network (Q.E.D). 

\end{proof}

\begin{proof}(Lemma~\ref{lem:composition})
Suppose that $\mathcal{M}_{i}$ is the composition of $l_{i}$ layers:
\begin{equation*}
    \begin{array}{ll}
        L^{(1)}_{i}: & \mathbb{R}^{n_{i}} \rightarrow \mathbb{R}^{n_{i}} \\
        \cdots & \\
        L^{(l_{i})}_{i}: & \mathbb{R}^{n_{l_{i}-1}} \rightarrow \mathbb{R}^{n_{i+1}} \\
    \end{array}
\end{equation*}

Then, we have the composition $\mathcal{M}_{1}, \cdots, \mathcal{M}_{k}$ is the composition of $\Sigma^{k}_{i=0} l_{i}$ layer as follows:
li\begin{equation*}
    \begin{array}{ll}
        L^{(1)}_{1}: & \mathbb{R}^{n_{1}} \rightarrow \mathbb{R}^{n_{1}} \\
        \cdots & \\
        L^{(l_{1})}_{1}: & \mathbb{R}^{n_{l_{1}-1}} \rightarrow \mathbb{R}^{n_{2}} \\
        \cdots & \\
        L^{(1)}_{k}: & \mathbb{R}^{n_{k}} \rightarrow \mathbb{R}^{n_{k}} \\
        \cdots & \\
        L^{(l_{k})}_{k}: & \mathbb{R}^{n_{l_{k}-1}} \rightarrow \mathbb{R}^{n_{k+1}} \\
    \end{array}
\end{equation*}

, where each layer $L^{(j)}_{i}$ is a feedforward layer in Def. ~\ref{def:ffnn_layer}

Therefore, we can conclude that the composition $\mathcal{M}_{1}, \cdots, \mathcal{M}_{k}$ is also a feed-forward neural network following Defintion~\ref{def:ffnn} (Q.E.D).
\end{proof}

\begin{proof} (Lemma~\ref{lem:agg_reduction})
Since the result of a GNN layer is produced by \[ \textbf{x}_i' = f_{upd} \left( \textbf{x}_i, f_{agg} \left( \left\{ f_{msg}( \mathbf{x}_{j}, \mathbf{x}_i, \mathbf{e}_{ji}) \forall j | (j, i) \in E  \right\} \right) \right) \]

Since the hidden features $\mathbf{x}_i \in \mathbb{R}^D$ is constructed by an FNN and $f_{msg}$ is an FNN, each $f_{msg}( \mathbf{x}_{j}, \mathbf{x}_i, \mathbf{e}_{ji})$ is also an FNN by Lemma~\ref{lem:composition}.
Furthermore, since $E$ remains unchanged, $f_{agg}$ is mean, max, or sum aggregation of these individual FNN is also an FNN. In detail, since $E$ is fixed, the number of incoming edges towards any node $i$ remains unchanged. Thus, every choice of $f_{agg}$ will lead to an FNN layer:
\begin{itemize}
    \item If $f_{agg}$ is mean aggregation, with a fixed number of elements, this reduces to a linear combination of its elements, with each coefficient set to a fixed value of inversed number of incoming edges.
    \item If $f_{agg}$ is max aggregation, with a fixed number of elements, this reduces to a fixed-size max pooling.
    \item If $f_{agg}$ is sum aggregation, with a fixed number of elements, this reduces to a linear combination of its elements with coefficients set to 1.
\end{itemize}
Thus, since $f_{agg}$ becomes an FNN layer that composites multiple other FNN layers as input (i.e., each $f_{msg}$), this part of a GNN layer is also an FNN. 
Finally, $f_{upd}$ is already FNN according to~\ref{def:ffnn}, and the GNN layer is reduced to an FNN (Q.E.D).
\end{proof}

Finally, we show that by recursively applying these reductions, a GNN is converted to an equivalent FNN in Theorem~\ref{thrm:equivalent}

\begin{proof} (Theorem~\ref{thrm:equivalent}) 
\noindent\textbf{Base Case}: We consider a GNN $\mathcal{M}$ with a single message-passing layer, as defined by Definition~\ref{def:message_passing_layer}. According to Lemma~\ref{lem:agg_reduction}, such a GNN layer can be reduced to an FNN layer because the operations of message passing can be emulated by appropriate weight matrices and non-linear function in an FNN. Thus for a GNN with a single layer, the theorem holds as the GNN is equivalent to an FNN.

\noindent\textbf{Inductive Hypothesis}: Assume that for a GNN $\mathcal{M}$ with $k$ layers, where $k \geq 1$, the computation of the first $k$ layers is equivalent to an FNN $\mathcal{M}_{F_k}$. This is our inductive hypothesis, which asserts that for $k$ layers, the computation on any graph structure $S$ can be performed by an equivalent FNN.

\noindent\textbf{Inductive Step}: Now, consider a GNN $\mathcal{M}$ with $k+1$ layers. By inductive hypothesis, the computation of the first $k$ layers of $\mathcal{M}$ can be performed by an FNN $\mathcal{M}_{F_k}$. For the $(k+1)$-th layer of $\mathcal{M}$, by Lemma~\ref{lem:agg_reduction}, this layer can also be reduced to an FNN layer.
By Lemma~\ref{lem:composition}, the composition of this additional layerwith the previously obtained FNN $\mathcal{M}_{F_k}$ yields another FNN, which we will denote as $\mathcal{M}_{F_{k+1}}$. Hence, the first $k$ layers combined with the $(k+1)$-th layer form an FNN equivalent to the original GNN with $k+1$ layers.

\noindent\textbf{Conclusion} Having shown that the addition of each layer in a GNN can be mirrored by a corresponding layer in an equivalent FNN, we conclude that for a GNN $\mathcal{M}$ with any number of layers, this procedure reduces $\mathcal{M}$ to an equivalent FNN $\mathcal{M}_F$ that computes the same function on any given fixed graph structure $S$. Therefore, Theorem~\ref{thrm:equivalent} is proven.

\end{proof}

\section{Transformation As Rewriting Rules} \label{app:transform}

Previously, we have shown that given a fixed structure (i.e., fixed edge set $E$). By reducing a $f_{agg}$ of the first layer, a GNN is reducible to an FNN in a bottom-up manner. Let us formalize these transformations as rewriting rules.

In detail, let's assume the fixed structure has the edge set $E$ with node set $V$. The input to GNN is now variable node feature matrix $\mathbf{X} \in \mathbb{R}^{|V| \times D_V}$ and edge feature matrix $\mathbf{E} \in \mathbb{R}^{|E| \times D_E}$.
We can further enforce an ordered edge set $E$ by introducing a comparison operator between 2 edge: $(i_1, j_1) \prec (i_2, j_2) \Leftrightarrow i_1 < i_2 \text{ or } i_1 = i_2 \land j_1 < j_2$. And let $idx(j, i)$ be the function that map from two nodes $(j, i)$ to the edge index in this order.
Recall that the set of all messages passing between all edges in $E$ is calculatable by $f_{msg}(\mathbf{x}_j, \mathbf{x}_i, \mathbf{e}_{ij}) \forall (j, i) \in E$. 
\begin{equation}
    \frac{idx(j, i) \mapsto idx_e}{f_{msg}(\mathbf{x}_j, \mathbf{x}_i, \mathbf{e}_{ij}) \forall (j, i) \in E \to f_{msg}(\mathbf{X}_{E_{idx_e, 0}, :}, \mathbf{X}_{E_{idx_e, 1}, :}, \mathbf{E}_{idx_e})}
\end{equation}
This has a meaningful implication: since the matrix-row selection is representable as a linear transformation, given the index function, we can now batch calculate all message with only linear operation.
Astute readers might notice that this is analogous to the adjacency matrix-based representation of GNN in~\cite{Kipf2016}. This is indeed a correct observation, however, we keep this annotation to keep precise semantic towards message passing networks~\cite{Gilmer2017}, which is a more general representation.
We can also calculate the index of incoming edges for each node $i$: $InIdx: i \mapsto \{idx(j, i) | (j, i) \in E\}$. 
Thus, for each node $i$, the aggregation can be rewritten as:
\begin{equation}
     \frac{\text{InIdx}(i) \mapsto IId_e}{f_{agg} \left( \left\{ f_{msg}( \textbf{x}_{j}, \textbf{x}_i, \textbf{e}_{ji}) \forall j | (j, i) \in E  \right\}\right) \to f_{agg} \left( f_{msg}(\mathbf{X}_{E_{IID_e, 0}, :}, \mathbf{X}_{E_{IID_e, 1}, :}, \mathbf{E}_{IID_e}) \right)}
\end{equation}
This $f_{agg}$ aggregates over a \textit{fixed-dimension} matrix by the result of $f_{msg}$, thus, it is further reducible based on the actual chosen aggregation function:
\begin{align}
   & \text{mean} \left( f_{msg}(\mathbf{X}_{E_{IID_e, 0}, :}, \mathbf{X}_{E_{IID_e, 1}, :}, \mathbf{E}_{IID_e}) \right) \to \frac{1}{|IID_e|} f_{msg}(\mathbf{X}_{E_{IID_e, 0}, :}, \mathbf{X}_{E_{IID_e, 1}, :}, \mathbf{E}_{IID_e}\\
    &\text{max} \left( f_{msg}(\mathbf{X}_{E_{IID_e, 0}, :}, \mathbf{X}_{E_{IID_e, 1}, :}, \mathbf{E}_{IID_e}) \right) \to \text{max}\left(f_{msg}(\mathbf{X}_{E_{IID_e, 0}, :}, \mathbf{X}_{E_{IID_e, 1}, :}, \mathbf{E}_{IID_e})\right)\\
    &\text{sum} \left( f_{msg}(\mathbf{X}_{E_{IID_e, 0}, :}, \mathbf{X}_{E_{IID_e, 1}, :}, \mathbf{E}_{IID_e}) \right) \to \underset{|IID_e|}{\left[1, 1, \ldots 1\right]}^\top\left(f_{msg}(\mathbf{X}_{E_{IID_e, 0}, :}, \mathbf{X}_{E_{IID_e, 1}, :}, \mathbf{E}_{IID_e})\right)
\end{align}

Finally, the update would remain the same but takes input as transformed $f_{msg}$ and $f_{agg}$ instead. This results in an FNN layer. By recursively applying these transformations, according to Theorem~\ref{thrm:equivalent}'s proof. We obtain an FNN.

\paragraph{Example}
For the case of $S_0$, $\mathbf{X} = \left( \begin{array}{l} s_0 \\ s_1 \end{array}\right) \in \mathbb{R}^{2 \times 1}$, and there would be no edge feature matrix $\mathbf{E}$. For the other example such as bellman-ford in \cite{Velickovic2019} where edge weight is needed, each edge feature matrix's row can store a single edge weight.
In the aforementioned example of BFS, the set of message would be from node $1$ to node $0$, node $0$ to node $0$ and node $1$ to node $1$.
Since it has 3 edges $\{(1, 0), (0, 0), (1, 1)\}$, for node $0$, $\text{InIdx}(0) = \{0, 1\}$ and $\text{InIdx}(1) = \{1\}$ for node $1$. 
We elaborate the process of transforming index selection to linear transformation as below.
We construct incoming edge selection for node $0$, $E_{s, 0} = \left(\begin{array}{ll} 1 & 0 \\ 0 & 1\end{array}\right)$ (first row selecting the index $0$ and 2nd row selecting the index $1$).
The message for node $0$ according to this selection matrix is: 
\begin{equation}
    f_{msg}\left(\mathbf{X} \cdot E_{s, 0}, \mathbf{X} \left[\begin{array}{ll} 1 & 0 \\ 1 & 0 \end{array}\right]\right) = \left[ \begin{array}{l} s_1 \\ s_0 \end{array}\right]
\end{equation}
The aggregated message for node $0$ would be transformed into: $\max(s_1, s_0)$. The same process is repeated for $s_1$.
Finally, transforming the update function would give us the final FNN:
\begin{equation}
	\mathbf{NS} = \mathbf{X}^{(1)} = \bar{\mathbf{M}} = \left(\begin{array}{l} \max(s_0, s_1) \\ s_1 \end{array}\right)
\end{equation}
% \section{Equivalence} \label{app:equiv}

% \begin{proof}(Theorem~\ref{thrm:equiv}
% Let $\mathcal{M}_{S}$ be a function representing the computation of GNN on the substructure $S$.

% From theorem~\ref{thrm:equivalent}, we have $\forall x: \mathcal{M}_{S} (x) \equiv \mathcal{M}_F(x)$. Hence, we have:
% \begin{equation*}
%     (pre(x) \Rightarrow post(\mathcal{M}_F(x))) \Leftrightarrow (pre(x) \Rightarrow post(\mathcal{M}_{S}(x)))
% \end{equation*}
% Therefore, a property $pre(x) \Rightarrow post(\mathcal{M}_F(x))$ is a feature property of the FNN $\mathcal{M}_F$ (in Theorem~\ref{thrm:equivalent}) iff the property is a feature property of  $\mathcal{M}_{S}$ (Q.E.D)
% \end{proof}

\section{Details of benchmark}\label{sec:benchmark_detailed}
Tabs.~\ref{tab:bfs_dynamic},~\ref{tab:dfs_dynamic}, and ~\ref{tab:blmfd_dynamic} shows the performance of individual GNNs and the respective proxy score of \tool{}.
Each of the three tables provides details about GNNs trained on BFS, DFS, and B-F problems. Columns \textbf{\#L} and \textbf{\#F} indicate the number of layers and hidden features in each layer, respectively, while \textbf{ltype} denotes the types of layers used, such as neural message passing or graph convolution, particularly for the BFS problem Fig.~\ref{fig:motiv_bfs}.

BFS GNNs delivered robust results, except for $\text{bfs}_1$ and $\text{bfs}_2$, with accuracy ranging from 0.9 to 1.0. DFS GNNs scored near-perfect accuracy for the \emph{next state} objective and high-performance ($\ge 0.95$) for the \emph{next visiting target} objective. B-F GNNs exhibited decent performance, with \tool{}'s properties achieving a proxy MSE below 0.012 for the next distance objective and a proxy accuracy from 0.89 for the next state objective.

\tool{}'s properties achieved perfect proxy accuracy for 15 GNNs and displayed high-confidence proxy accuracy from 0.888 to 0.918 for the remaining 4 GNNs. However, perfect proxy scores were only achieved for $\text{dfs}_5$ in \emph{next state} and $\text{dfs}_1$ in \emph{next target}.

\tool{}'s properties showed high proxy scores for both objectives, ranging from 0.93 for 8 GNNs to 0.79 for $\text{dfs}_8$ in \emph{next state}, and from 0.8799 for the \emph{next visiting target} objective.

Looking at the \textbf{IR} values across tables \ref{tab:bfs_dynamic}, \ref{tab:dfs_dynamic}, and \ref{tab:blmfd_dynamic}, we see that for most cases in BFS and DFS's next state and DFS's next visiting target, the addition of dynamic predicates leads to an improvement of less than 20\%.  This indicates that the structure-specific properties can largely model the GNNs' behaviors even without dynamic predicates, but these dynamic features still play a role in enhancing the precision of the properties.
For certain cases like $\text{dfs}_8$ where structure-specific properties fail to adequately capture GNN behavior, the dynamic analysis can provide the necessary conditions for the property to hold in the full graph. This is seen in sections Section~\ref{rq:reference} and Section~\ref{sec:backdoor}.

Interestingly, dynamic analysis drastically reduces proxy MSE for B-F's next distance by factors between $10^3$ to $10^6$. Moreover, the improvement of B-F's next state objective is significantly higher than its BFS counterpart despite having the same reference GNN. 

Overall, dynamic analysis can boost the accuracy of the final properties for most cases and in some cases, it can even be the main contributor to the property's precision in modeling the behavior of GNNs.
In summary, the inferred properties demonstrate strong proxy measures across all three problems, indicating \tool{}'s capability to identify properties that effectively represent the GNNs' behaviors in both classification and regression tasks. It's important to note that there are instances where the model performance is less robust yet the proxy accuracy remains highly confident. Conversely, there are also cases of perfect model performance accompanied by less-than-perfect proxy accuracy, which we have discussed in Section~\ref{sec:discussion}.
\begin{table}
   \centering
   \caption{Performance of 17 trained GNN for BFS.}
   \small
   \begin{tabular}{|l|c|c|c|c|c|c|}
        \hline
        \textbf{Name} & \textbf{\#L/\#F} & \textbf{ltype} & \textbf{SA} & \textbf{PA} & \textbf{IR} \\
        \hline
        ref & 1/1 & mpnn & 1 & 1 & 0.115 \\
        $\text{bfs}_1$ & 1/2 & gcn & 0.73 & 1 & 0.053 \\
        $\text{bfs}_2$ & 1/2 & gcn & 1& 1 & 0.144 \\
        $\text{bfs}_3$ & 1/4 & gcn & 1 & 1 & 0.144 \\
        $\text{bfs}_4$ & 2/2 & gcn & 0.63 & 1 & 0 \\
        $\text{bfs}_5$ & 2/4 & gcn & 0.90 & 0.888 & 0.0042 \\
        $\text{bfs}_6$ & 2/8 & gcn & 0.90 & 0.888 & 0.0033 \\
        $\text{bfs}_7$ & 3/2 & gcn & 1 & 1 & 0.115 \\
        $\text{bfs}_8$ & 3/4 & gcn & 1 & 0.89 & 0.0011 \\
        $\text{bfs}_9$ & 3/8 & gcn & 1 & 1 & 0.115 \\
        $\text{bfs}_{10}$ & 1/2 & mpnn & 1 & 1 & 0.115 \\
        $\text{bfs}_{11}$ & 1/4 & mpnn & 1& 1 & 0.115 \\
        $\text{bfs}_{12}$ & 2/2 & mpnn & 1& 1 & 0.115 \\
        $\text{bfs}_{13}$ & 2/4 & mpnn & 1& 1 & 0.115 \\
        $\text{bfs}_{14}$ & 2/8 & mpnn & 1& 1 & 0.115 \\
        $\text{bfs}_{15}$ & 3/2 & mpnn & 1& 0.918 & 0.0026 \\
        $\text{bfs}_{16}$ & 3/4 & mpnn & 1& 1 & 0.115 \\
        $\text{bfs}_{17}$ & 3/8 & mpnn & 1& 1 & 0.115 \\
     \hline
    \end{tabular}
    \label{tab:bfs_dynamic}
\end{table}

\begin{table}
    \centering
    \caption{Performance of 8 trained GNNs and reference GNN accuracy for DFS.}
    \small
    \begin{tabular}{|c|c|c|c|c|c|c|c|}
        \hline
        \multirow{2}{*}{\textbf{Name}} & \multirow{2}{*}{\textbf{\#L/\#F}} & \multicolumn{3}{c|}{\textbf{Next State}} & \multicolumn{3}{c|}{\textbf{Next Visiting Target}}\\
        \cline{3-8}
        &  & \textbf{SA} & \textbf{SPA} & \textbf{SIR} & \textbf{TA} & \textbf{TPA} & \textbf{TIR}\\
        \hline
        ref & 2/2 & 1 & 0.931 & 0.084 & 1 & 0.880 & 0.150 \\
        $\text{dfs}_1$ & 1/2 & 1 & 0.955 & 0.402 & 0.950 & 1 & 0 \\
        $\text{dfs}_2$ & 1/4 & 1 & 0.955 & 0.010 & 0.964 & 0.959 & 0.009 \\
        $\text{dfs}_3$ & 2/2 & 1 & 0.956 & 0.026 & 0.961 & 0.963 & 0.001 \\
        $\text{dfs}_4$ & 2/4 & 1 & 0.972 & 0.164 & 0.988 & 0.889 & 0.167 \\
        $\text{dfs}_5$ & 2/8 & 1 & 1 & 0 & 0.995 & 0.889 & 0.061 \\
        $\text{dfs}_6$ & 3/2 & 0.990 & 0.994 & 0.551 & 0.960 & 0.993 & 0.004 \\
        $\text{dfs}_7$ & 3/4 & 1 & 0.993 & 0 & 0.988 & 0.931 & 0.091 \\
        $\text{dfs}_8$ & 3/8 & 1 & 0.791 & 1.667 & 0.996 & 0.899 & 0.064 \\
        \hline
    \end{tabular}
    \label{tab:dfs_dynamic}
\end{table}

\begin{table}
   \centering
   \caption{Performance of 8 trained GNNs and reference GNN accuracy for B-F.}
   \small
   \begin{tabular}{|c|c|c|c|c|c|c|c|}
        \hline
        \multirow{2}{*}{\textbf{Name}} & \multirow{2}{*}{\textbf{\#L/\#F}} & \multicolumn{3}{c|}{\textbf{Next Distance}} & \multicolumn{3}{c|}{\textbf{Next State}} \\
        \cline{3-8}
        &  & \textbf{DE} & \textbf{DPE} & \textbf{DIR} & \textbf{SA} & \textbf{SPA} & \textbf{SIR} \\
        \hline
        ref & 1/2 & 0.0 & $2e^{-9}$ & 1 & 1 & 1 & 0.145 \\
        $\text{blmfd}_{1}$ & 1/2 & $1.4e^{-4}$ & $1.04e^{-3}$ & 0.996 & 1 & 1 & 0.364 \\
        $\text{blmfd}_{2}$ & 1/4 & $1.2e^{-4}$ & $5.21e^{-3}$ & 0.999 & 1 & 0.927 & 0.162 \\
        $\text{blmfd}_{3}$ & 2/2 & $3e^{-3}$ & $8.32e^{-3}$ & 0.998 & 1 & 0.891 & $5.5e^{-4}$ \\
        $\text{blmfd}_{4}$ & 2/4 & $1e^{-3}$ & $5.2e^{-4}$ & 1 & 1 & 1 & 0.325 \\
        $\text{blmfd}_{5}$ & 2/8 & $1.9e^{-4}$ & $2.3e^{-4}$ & 1 & 1 & 1 & 0.42 \\
        $\text{blmfd}_{6}$ & 3/2 & $1.8e^{-4}$ & $3.01e^{-3}$ & 0.999 & 1 & 0.911 & 0.697 \\
        $\text{blmfd}_{7}$ & 3/4 & $1.23e^{-4}$ & $7.93e^{-7}$ & 1 & 1 & 1 & 0.551 \\
        $\text{blmfd}_{8}$ & 3/8 & $1.7e^{-4}$ & $1.15e^{-2}$ & 0.999 & 1 & 1 & 0.535 \\
     \hline
    \end{tabular}
    \label{tab:blmfd_dynamic}
\end{table}

\end{document}